\pdfoutput=1

\documentclass[11pt]{article}

\usepackage{latex/acl}
\newcommand{\emark}{\ding{229}}
\usepackage{times}
\usepackage{latexsym}
\usepackage{marvosym}

\usepackage[T1]{fontenc}

\usepackage[utf8]{inputenc}

\usepackage{microtype}

\usepackage{inconsolata}

\usepackage{hyperref}
\usepackage{array,multirow,graphicx}
\usepackage{rotating}
\usepackage{csquotes}
\usepackage{tabularx}
\usepackage{graphicx}
\usepackage{kotex}
\usepackage{makecell}
\usepackage{colortbl}

\usepackage{amssymb}
\usepackage{pifont}
\usepackage{arydshln} 
\usepackage[caption=false]{subfig}

\usepackage{amsmath}
\usepackage{amsfonts}
\usepackage{amssymb}

\usepackage{caption}
\usepackage{csquotes}

\DeclareCaptionType{equ}[][]

\usepackage[normalem]{ulem}
\usepackage{hhline}
\usepackage{booktabs}

\usepackage{soul}
\usepackage{xcolor}
\usepackage{tcolorbox}
\sethlcolor{green}

\sethlcolor{yellow}

\sethlcolor{gray}

\usepackage{dsfont}
\usepackage{stmaryrd}
\usepackage{float}
\usepackage{tikz}

\definecolor{right}{RGB}{0,128,96}
\definecolor{wrong}{RGB}{192,0,32}

\definecolor{green}{RGB}{59, 125, 35}
\definecolor{skyblue}{RGB}{33, 95, 154}
\definecolor{orange}{RGB}{255, 102, 0}
\definecolor{pink}{HTML}{C8635B}
\definecolor{softsky}{HTML}{58ACB3}

\def \mydata{\textsc{CCPT}}

\newcommand{\Green}[1]{\textcolor{green}{#1}}
\newcommand{\Orange}[1]{\textcolor{orange}{#1}}
\newcommand{\Blue}[1]{\textcolor{skyblue}{#1}}
\newcommand{\Right}[1]{\textcolor{right}{#1}}
\newcommand{\Wrong}[1]{\textcolor{wrong}{#1}}
\newcommand{\cmark}{\Right{\ding{51}}}
\newcommand{\tmark}{{\textcolor{gray}{$\Delta$}}}

\newcommand{\xmark}{\Wrong{\ding{55}}}

\newcommand{\githuburl}[0]{\url{https://github.com/seokwon99/CCPT.git}}

\usepackage{booktabs,arydshln}

\makeatletter
\def\adl@drawiv#1#2#3{%
        \hskip.5\tabcolsep
        \xleaders#3{#2.5\@tempdimb #1{1}#2.5\@tempdimb}%
                #2\z@ plus1fil minus1fil\relax
        \hskip.5\tabcolsep}
\newcommand{\cdashlinelr}[1]{%
  \noalign{\vskip\aboverulesep
           \global\let\@dashdrawstore\adl@draw
           \global\let\adl@draw\adl@drawiv}
  \cdashline{#1}
  \noalign{\global\let\adl@draw\@dashdrawstore
           \vskip\belowrulesep}}
\makeatother

\usepackage{graphicx}
\usepackage{longtable}
\usepackage[ruled, vlined]{algorithm2e}

\SetCommentSty{mycommfont}

\title{\textit{Is a Peeled Apple Still Red?} Evaluating LLMs' Ability for Conceptual Combination with Property Type}

\author{
\quad \textbf{Seokwon Song}$^{\star, 1}$
\quad \textbf{Taehyun Lee}$^{\star, 1}$
\quad \textbf{Jaewoo Ahn}$^{1}$ \\ 
\quad \textbf{Jae Hyuk Sung}$^{2, \sharp}$
\quad \textbf{Gunhee Kim}$^{1}$\\
$^1$Seoul National University\quad $^2$Korea University \\
\texttt{\small \{seokwon.song, taehyun.lee, jaewoo.ahn\}@vision.snu.ac.kr, \small okaybody10@korea.ac.kr} \\
\texttt{\small gunhee@snu.ac.kr} \\
}

\newcommand{\correspondingfootnote}{
    \let\oldthefootnote=\thefootnote
    \renewcommand{\thefootnote}{}
    \footnotetext{$\star$ Authors equally contributed.}
    \footnotetext{$\sharp$ Work done during internship at Seoul National University.}
    \let\thefootnote=\oldthefootnote
}

\begin{document}

\maketitle
\begin{abstract}

\correspondingfootnote
Conceptual combination is a cognitive process that merges basic concepts, enabling the creation of complex expressions. During this process, the properties of combination (e.g., the whiteness of a peeled apple) can be inherited from basic concepts, newly emerge, or be canceled. However, previous studies have evaluated a limited set of properties and have not examined the generative process.
To address this gap, we introduce the Conceptual Combination with Property Type dataset (\mydata), which consists of 12.3K annotated triplets of noun phrases, properties, and property types. Using \mydata, we establish three types of tasks to evaluate LLMs for conceptual combination thoroughly.
Our key findings are threefold:
(1) Our automatic metric grading property emergence and cancellation closely corresponds with human judgments.
(2) LLMs, including OpenAI's o1, struggle to generate noun phrases which possess given emergent properties.
(3) Our proposed method, inspired by cognitive psychology model that explains how relationships between concepts are formed, improves performances in all generative tasks.
The dataset and experimental code are available at \githuburl{}.

\end{abstract}

\section{Introduction}

The conceptual combination is a fundamental cognitive process that synthesizes multiple basic concepts into a novel concept~\cite{wisniewski1997concepts, thagard1984conceptual, coutanche2019conceptual, ward2001creative}. Combining concepts is potentially limitless and plays an important role in various fields such as engineering, science~\cite{hampton1997emergent}, figurative language and literature~\cite{ward2001creative}. For example, `a wilted flower' evokes our sadness because we internally compare a flower's former beauty and current feeble state. A skilled writer knows it and kindles diverse imagery through conceptual combinations. 
The potential for such conceptual combinations is vast, so large language models (LLMs) should be able to interpret or generate meaningful insights even for rare or novel combinations.
\begin{figure}[t!]
\centering
\includegraphics[width=\linewidth]{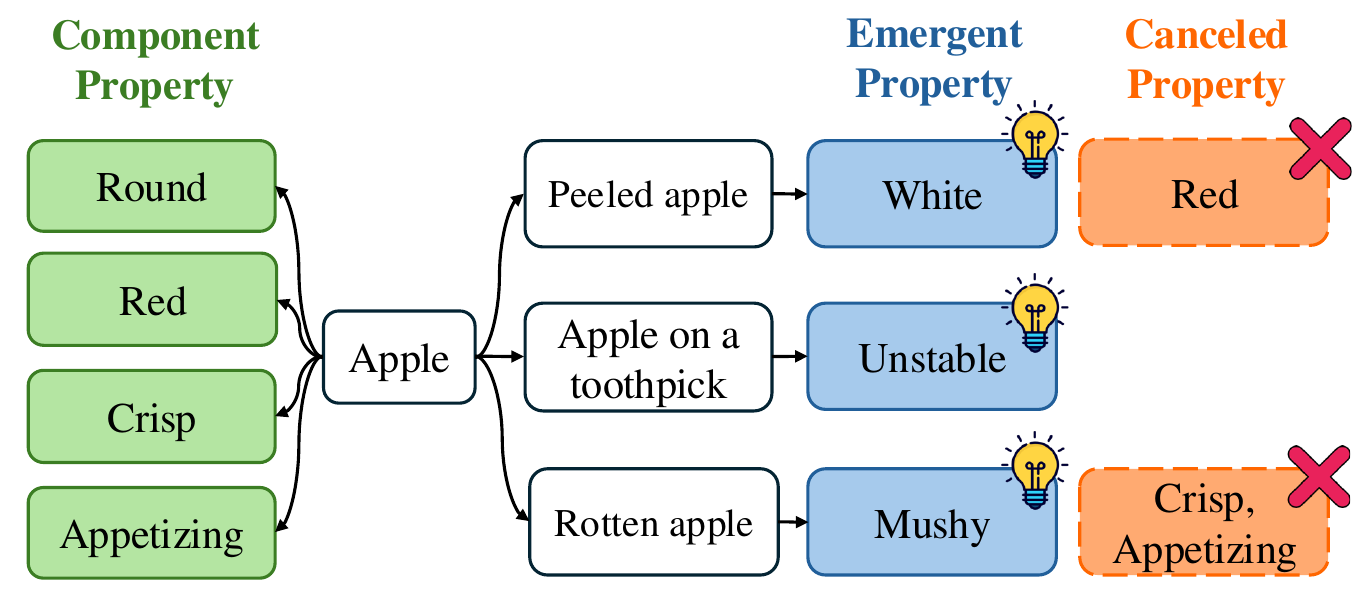}
\caption{Three types of properties derived from conceptual combination with an example of ``apple''. Different concepts are formed by adding other concepts to ``apple''. The \Green{green} properties are \textit{component properties} of the basic concept ``apple''. The \Blue{blue} and \Orange{orange} are \textit{emergent} and \textit{canceled properties}, respectively.}
\label{fig:intro_figure}
\vspace{-3mm}
\end{figure}

Several works have explored concept knowledge of language models, as outlined in Table \ref{tab:benchmarks}. Basic Concept Probing examines whether language models understand individual concepts at the entity (e.g., ``horses are mammals'') and event levels (e.g., ``drinking tea leads to feeling refreshed''). However, this approach mainly focuses on single concepts rather than combinations. Noun Compound Interpretation studies how models generate plausible meanings for combinations of nouns (e.g., ``dog house'' means ``a house for a dog''), but does not explore deeper properties of combinations. In conceptual combination works, Big-Bench-CC investigates emergent properties. FakeReef delves into a special type of canceled property (membership inference). Still, none of them addresses all of emergent, component, or canceled properties nor generative tasks related to creating combinations or property. 

It is essential to understand not only properties that emerge from a combination, but also whether existing properties maintain or disappear~\cite{springer1992featureavailability, gagne2017conceptual, estes2020relevance}. Following \citet{springer1992featureavailability}, we identify three types of properties in conceptual combination, which are illustrated using the example of an ``apple'' in Figure~\ref{fig:intro_figure}: (1) \textit{Component property} refers to a property inherent to individual concepts (e.g., ``red'' for ``apple on a toothpick''), (2) \textit{Emergent property} is newly from the combination of concepts (e.g., ``unstable'' for ``apple on a toothpick''), and (3) \textit{Canceled property} is negated due to the combination (e.g., ``crisp'' for ``rotten apple''). These three types form a complete set of properties related to conceptual combination since they indicate \textit{already present}, \textit{newly created}, and \textit{newly disappeared}.

\interfootnotelinepenalty=10000
\newcolumntype{?}{!{\vrule width 1pt}}

\begin{table}[t!]
\small
\centering\resizebox{\linewidth}{!}{
\begin{tabular}{llcccc}
\toprule[1.2pt]
\multirow{2}{*}{\makecell[l]{\textsc{dataset}}} & \multirow{2}{*}{\makecell[l]{\# \textsc{Q}}} & \multicolumn{3}{c}{\textsc{Property}}& \multirow{2}{*}{\makecell{\textsc{MC}}} \\
\cline{3-5}
  & & \textsc{com} & \textsc{eme} & \textsc{can} &\\ \midrule
\multicolumn{6}{l}{\textbf{\textit{Basic Concept Probing}}}\\
\makecell[l]{BERT-IS-A (\citeyear{lin2022does})}& 11,868 & \cmark  & \xmark & \xmark & \xmark  \\
\makecell[l]{COPEN (\citeyear{peng2022copen})} & 10,624 & \cmark  & \xmark & \xmark & \xmark  \\
\makecell[l]{PLM-Ontology (\citeyear{wu2023plms})}& 9,687 & \cmark  & \xmark & \xmark & \xmark  \\
\midrule
\multicolumn{6}{l}{\textbf{\textit{Noun Compound Interpretation}}}\\
\makecell[l]{SemEval-2013-Task4 (\citeyear{hendrickx2013semeval})}& 355& \xmark  & \xmark & \xmark & \cmark  \\
ChocolateBunny (\citeyear{coil2023chocolate})& 105& \xmark  & \xmark & \xmark & \cmark  \\
COMPUN (\citeyear{rambelli2024can})  & 918 & \xmark  & \xmark & \xmark & \cmark  \\\midrule
\multicolumn{6}{l}{\textbf{\textit{Conceptual Combination}}}\\
\makecell[l]{BIG-bench-CC (\citeyear{srivastava2023bigbench})}& 103& \tmark  & \cmark & \xmark & \cmark  \\
FakeReef (\citeyear{ross2024artificial}) & 854 & \cmark & \xmark & \cmark & \cmark \\
\mydata~(ours)& 12,315& \cmark  & \cmark & \cmark & \cmark  \\ \bottomrule[1.2pt]
\end{tabular}
}

\caption{Comparison of our dataset with previous concept property probing and noun compound interpretation datasets. Each column denotes as follows. \# Q: the number of questions with gold answers. \textsc{com}, \textsc{eme}, and \textsc{can} denotes component, emergent and canceled property, respectively. \textsc{MC}: whether the combinations of multiple concepts are considered. In Big-bench-CC, \tmark~indicates that component properties are present in a dataset as wrong options, but it does not specify what properties are component properties.}
\label{tab:benchmarks}
\vspace{-5mm}
\end{table}

In this study, we aim to fill this gap by proposing the \textbf{C}onceptual \textbf{C}ombination with \textbf{P}roperty \textbf{T}ype (\textbf{\mydata}) dataset. Our dataset contains 12.3K annotated triplets of noun phrase comprised of two basic concepts, property and property type. The basic concepts in noun phrases are comprised of 9K unique uni-gram concepts from ConceptNet~\cite{speer2017conceptnet}.

To evaluate LLMs' ability on conceptual combination, we introduce two generative and one classification tasks, detailed in Table \ref{tab:task_sample} and \S \ref{sec:taskform}.
Based on the prior works in cognitive psychology and linguistics, each task is designed to evaluate cognitive processes related to conceptual combination as follows. (1) \textit{Property induction}: imagine the property of noun phrase~\cite{wilkenfeld2001similarity, estes2002emergence}, (2) \textit{Noun phrase completion}: create a noun phrase to represent certain property, highlighted as a critical point of creativity~\cite{kohn2011conceptual}, and (3) \textit{Property type prediction}: predict a type of property to identify the origin of property~\cite{gagne2017conceptual}.

Extensive experiments show that: (1) our automatic evaluation metric, based on the LLM-as-a-judge~\cite{zheng2023judging} method to detect property emergence and cancellation scores, closely matches human evaluations; (2) current LLMs, including o1~\cite{openai2024o1}, struggle to create noun phrases with truly emergent properties, often defaulting to results where one component already has the target property; and (3) our proposed method, motivated by the spread activation model~\cite{anderson1983spreading, collins1975spreading} in cognitive psychology, improves performance across all generative tasks, indicating that considering diverse relationships between concepts enhances conceptual combination.

\section{Related Work}
\subsection{Conceptual Combination}
Conceptual combination is a mental process of combining concepts. Unique features often arise from the combination, especially when the combinations of elements are unusual~\cite{estes2002emergence, ward2007creative}.
People come up with novel ideas to merge conflicts; thus, it is a prominent part of creativity~\cite{hampton1997emergent}.
In scientific discovery, conceptual combination creates new scientific concepts~\cite{thagard1984conceptual}, such as ``light wave.'' Even if nobody could observe light waves, the properties shared with waves, such as reflection, led to a theory that light is a wave, leading to much progress in science.

\subsection{Conceptual Combination in NLP} 
In NLP, understanding conceptual combination has been studied in two different lines. For instance, interpreting ``pet bird'' as ``a bird kept as a pet'' exemplifies \textit{noun compound interpretation}. Alternatively, imagining ``pet bird''s attributes like ``probably lives in a cage'' illustrates \textit{property verification}.

Noun Compound Interpretation focuses on generating plausible meanings for noun-noun compounds, such as interpreting ``dog house'' as ``a house for a dog''. The goal is to resolve ambiguity and provide clear interpretations of compound nouns. While related to combining nouns, it mainly emphasizes understanding the relationships between the words~\cite{hendrickx2013semeval,shwartz2018olive,shwartz2019still, coil2023chocolate, rambelli2024can}.

The second line of research focuses on the properties of conceptual combinations. \citet{srivastava2023bigbench} introduced a benchmark for identifying emergent properties in conceptual combinations. However, their approach has limitations: (1) it only identifies emergent properties, (2) the benchmark is limited to multiple-choice selection tasks without generative tasks, and (3) it does not involve creating conceptual combinations. In contrast, our approach (1) uses property types as part of the constraints and (2) includes both generative tasks and multiple-choice selection tasks. Concurrent work~\cite{ross2024artificial} deeply explores LLM's understanding of modifier adjective-noun conceptual combination with a membership inference question such as `Is a fake reef still a reef?'. The work shares interest with us about property cancellation. However, there's still a gap in evaluating LLMs for conceptual combination in generative way.

\section{The \mydata~Benchmark}

\subsection{Task Formulation}
\label{sec:taskform}
\begin{table}[t!]
\centering
\small
\begin{tabular}{p{0.15\linewidth}p{0.73\linewidth}} 
\toprule[1.2pt]
\multicolumn{2}{c}{\textsc{\textbf{Task 1: Property Induction}}} \\ 
\cmidrule[\heavyrulewidth]{1-2} 
\textbf{Inst.} &
Given a noun phrase and a property type, generate a \textul{property} that satisfies both. \\[2pt]\cdashlinelr{1-2}
\textbf{Input} & Noun phrase: An apple on a toothpick\newline Type: Emergent \\
\cmidrule{1-2} 
\textbf{Response1}& Unstable\\ 
\cdashlinelr{1-2} 
 &\emark~\textit{``Apple on a toothpick'' is unstable, but `apple' or `toothpick' is seldom related to it: \Right{good output (7/10)}} \\
 \cmidrule{1-2}
\textbf{Response2} & Delicious \\
\cdashlinelr{1-2} 
&\emark~\textit{``Apple'' is already delicious, and that makes `apple on a tootkpick' also delicious: \Wrong{bad output (2/10)}} \\
\midrule[1.5pt]
\multicolumn{2}{c}{\textsc{\textbf{Task 2: Noun Phrase Completion}}} \\ 
\cmidrule[\heavyrulewidth]{1-2} 
\textbf{Inst.} &
Given a head noun, a property and a type, generate a \textul{noun phrase} that satisfies both. \\[2pt]\cdashlinelr{1-2}
\textbf{Input} & Head noun: Apple\newline Property: Rare\newline Type: Emergent \\
\cmidrule{1-2} 
\textbf{Response1}& A blue apple\\ 
\cdashlinelr{1-2} 
 &\emark~\textit{``A blue apple'' has new emergent property `rare' even if `apple' is common and `blue' is unrelated to `rare': \Right{good output (8/10)}} \\
 \cmidrule{1-2}
\textbf{Response2} & ``An overripe apple''  \\
\cdashlinelr{1-2} 
&\emark~\textit{``An overripe apple'' is not considered rare at all: \Wrong{bad output (1/10)}} \\
\midrule[1.5pt]
\multicolumn{2}{c}{\textsc{\textbf{Task 3: Type Prediction}}} \\ 
\cmidrule[\heavyrulewidth]{1-2} 
\textbf{Inst.} &
Given a noun phrase and a property, choose a \textul{type of property}. \\
 &\vspace{-6pt}
(1) Emergent \hspace{1.5pt}(2) Canceled \hspace{1.5pt} (3) Component \newline(4) Others \\[2pt]\cdashlinelr{1-2}
\textbf{Input} & Noun phrase: A green apple\newline Property: Good for health \\
\cmidrule{1-2} 
\textbf{Response1}& (3) Component\\ 
\cdashlinelr{1-2} 
 &\emark~\textit{``Component property'' is a \Right{good output}, because `apple' is already good for health.} \\
\bottomrule[1.2pt]
\end{tabular}
\caption{
    Each example consists of a task instruction, input, and responses with their grading (Tasks 1 and 2) and correctness (Task 3). The explanations \emark~in italics are added by the authors for clarification. Note that LLMs are evaluated solely based on their responses.
}
\label{tab:task_sample}
\vspace{-5mm}
\end{table}

Before describing our task formulation, we begin with its key elements: $\{\mathcal{N}(\mathcal{H}, \mathcal{M}), \mathcal{P}, \mathcal{T}\}$, with two examples of ``peeled apple'' and ``egg in the batter'' for better understanding.
\setlength{\itemsep}{0pt} 
\setlength{\parskip}{0pt} 
\setlength{\parsep}{0pt}  
\begin{itemize}
    \item \textbf{Noun phrase} $\mathcal{N}$ is noun phrase comprised of two concepts (e.g., ``peeled apple'' and   ``egg in the batter'').
    \vspace{-1mm}
    \item \textbf{Head noun} $\mathcal{H}$ provides the central meaning of the combination $\mathcal{N}$ (e.g., ``apple'' and   ``egg'').
    \item \textbf{Modifier} $\mathcal{M}$ adds additional meaning to the head noun $\mathcal{H}$ (e.g., ``peeled'' and ``batter'').
    \item  \textbf{Property} $\mathcal{P}$ is attributed to the combination $\mathcal{C}$ (e.g., ``white'' for a peeled apple; ``nutritious'' for the egg in the batter).
    \item \textbf{Type of property} $\mathcal{T}$ introduces the origin of the property $\mathcal{P}$ (e.g., ``emergent property'' - ``white'' for a peeled apple, ``component property'' - ``nutritious'' for the egg in the batter) 
\end{itemize}

Based on the above configuration, we devise three types of tasks by hiding one configuration and using the remaining configurations to predict it. This task formulation includes two generative tasks: property induction and noun phrase completion, and one classification task: property type prediction.

\subsubsection{Property Induction~\small{($\mathcal{N},  \mathcal{T} \rightarrow \mathcal{P}$)}}
\label{induction}

The interpretation of novel combinations by listing property has been a well-explored area in previous research on human cognition~\cite{wilkenfeld2001similarity, estes2002emergence}, with a significant focus on how people comprehend novel expressions in sentences and discourse~\cite{swinney2007conceptual}.
In this task, LLMs are instructed to identify the properties of combinations that align with the given property types (emergent, canceled property).

For example, in Table \ref{tab:task_sample} (top), LLMs can identify an emergent property by finding a property that is not present in the individual concept but emerges in the noun phrase, such as ``unstable'' for ``apple on a toothpick''.

\subsubsection{Noun Phrase Completion \small{($\mathcal{H}, \mathcal{P}, \mathcal{T} \rightarrow \mathcal{N}$)}}
\label{completion}
Generating new concepts by combining existing ones is key to creativity ~\cite{kohn2011conceptual}. In this task, LLMs generate noun phrases by adding modifiers to head nouns to represent emergent or canceled properties. For emergent properties, the modifier should not imply the property on its own, but the combination should. In contrast, for canceled properties, the modifier effectively negates the head noun’s property.

For example, in Table \ref{tab:task_sample} (middle), to represent the emergent property ``rare'' with the head noun ``apple'', LLMs may consider a noun phrase like ``blue apple''. The modifier ``blue'' does not directly suggest the property ``rare'', but when combined with ``apple'', it possesses the given property.

\subsubsection{Property Type Prediction~\small{($\mathcal{N}, \mathcal{P} \rightarrow \mathcal{T}$)}}
\label{prediction}
Understanding how combined concepts gain or lose certain properties is an essential process for concept theories~\cite{gagne2017conceptual}.

In this task, LLMs identify how a property relates to a noun phrase. For example, in Table \ref{tab:task_sample}, the property ``good for health'' in ``a green apple'' is an \textit{component property}, as the apple is already good for health itself.

\subsection{Data Collection}
\label{data construction}
\begin{figure*}[t!]
\centering
\includegraphics[width=0.95\textwidth]{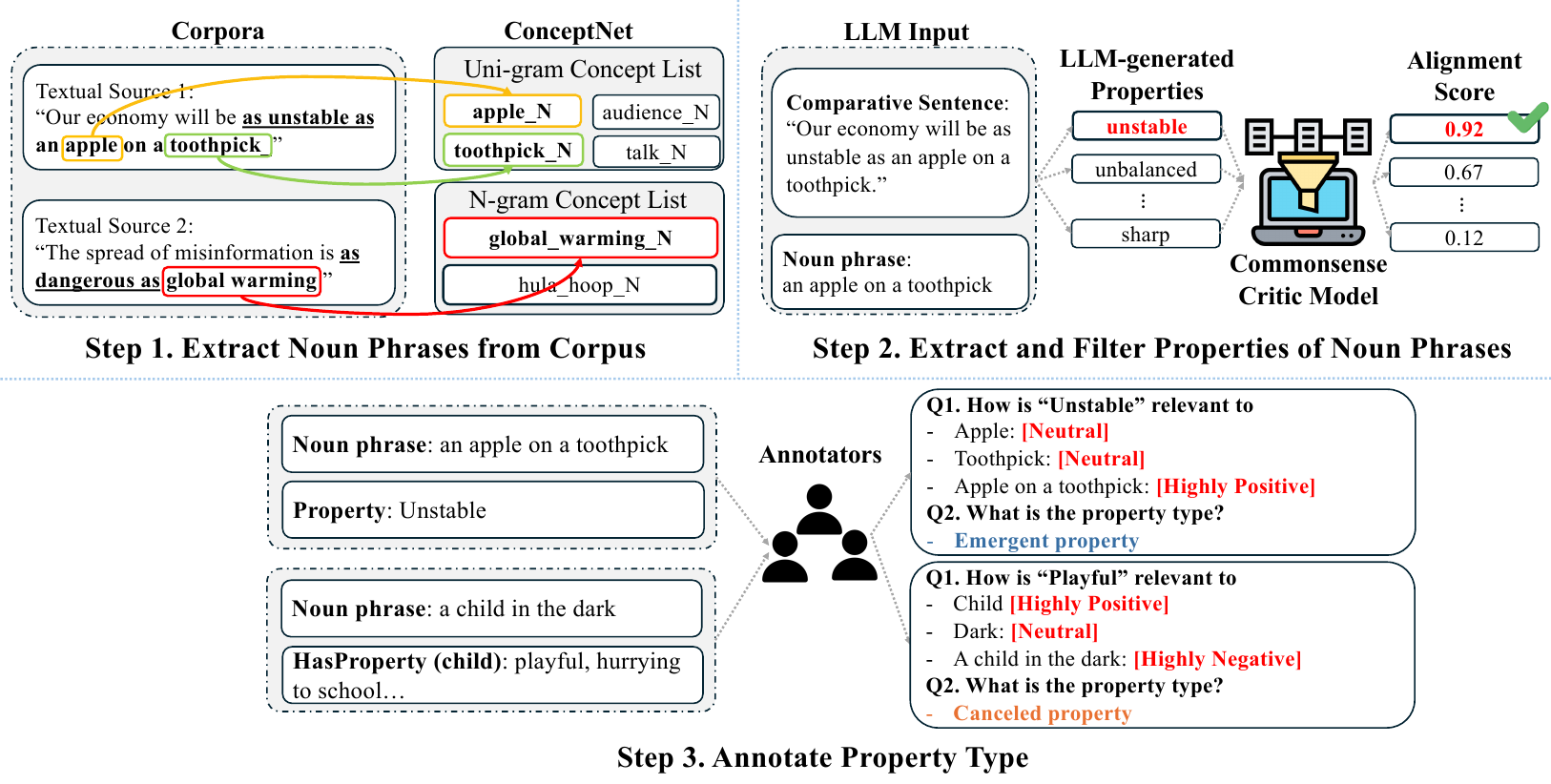}
\caption{Overview of our data collection pipeline for conceptual combination through automated and human-driven data annotation.}
\label{fig:pipeline}
\vspace{-5mm}
\end{figure*}

We propose the dataset, \textbf{C}onceptual \textbf{C}ombination with \textbf{P}roperty \textbf{T}ype (CCPT), to address the three tasks in \S~\ref{sec:taskform}. We both use a automated methods and human filtering. Detailed sources of textual corpora are described in Appendix \ref{textual_corpora}.

\noindent\textbf{Step 1. Extract and Filter Combinations.} 

First, we extract noun phrases from the corpus. To obtain noun phrases with a property hint in a sentence, we extract sentences that contain ``like'' or ``as'' which compare one concept to another. These comparisons explicitly highlight the characteristics (e.g., ``our economy will be as \textul{unstable} as an apple on a toothpick''). This yields 51.0M comparative sentences—8\% of the paragraph in the original corpus contained such sentences.

To avoid common expressions, such as proper nouns or idioms, whose meanings can be memorized from training corpus, we exclude any N-grams found in ConceptNet.
We utilize uni-gram concept set from ConceptNet and use it as basic concepts. In the end, we collected 136.0K comparative sentences containing combination made up of two uni-gram concepts.

\noindent\textbf{Step 2. Extract and Filter Properties.}

\textbf{Emergent \& Component Candidates.} 
Property extraction by syntactic patterns such as ``ADJ/ADV like C'' or ``as ADJ/ADV as C'' often misses implicit properties (e.g., ``the storm was almost like a raging bull'') or multi-word properties (e.g., ``they \textit{crashed together} like a boat on the rocks''). 

To address this, we use GPT-4o-mini~\cite{openai2024gpt4omini} to extract 10 properties from given comparative sentence for each combination. Then VERA-T5-XXL~\cite{liu2023vera} filters out unlikely properties (with an alignment score under 0.7), resulting in 41.6K noun phrases and 211.0K properties.

Then we extract candidates for each property type. For emergent properties, with GPT-4o-mini, we select instances where neither the head noun $\mathcal{H}$ nor the modifier $\mathcal{M}$ already possess the property $\mathcal{P}$ of the noun phrase $\mathcal{N}$. We then limit the selection to at most five instances per noun phrase, resulting in 3,851 candidates. For component properties, we randomly sample 10K candidates.

\textbf{Canceled Candidates.}
Since we cannot gather canceled properties from the corpus, we use two additional sources to collect the properties of head nouns: one for the ``HasProperty'' relations from ConceptNet and another from GPT-4o-mini, following Step 2.  
We randomly sample 2K noun phrases and gather up to 10 properties for each head noun from each source. Then, VERA-T5-XXL filters out the most likely properties (with an alignment score above 0.7), resulting in 2K noun phrases and a total of 23K properties.

With GPT-4o-mini, we select instances where the noun phrase $\mathcal{N}$ no longer possesses the property $\mathcal{P}$ of the head noun $\mathcal{H}$. Then, we select at most 5 data with the same noun phrase. We then limit the selection to at most five instances per noun phrase, resulting in 4,457 candidates.

\noindent\textbf{Step 3. Annotate Property Types.}
The annotation process consists of three stages: first, assign a 5-point Likert scale of relevance score ranging from Highly Negative to Highly Positive; second, annotate the property type; third, find the toxicity.

In total, we have 12,315 data points, the type of property $\mathcal{T}$ consisting of 2501 emergent properties, 1613 canceled properties, and 8201 component properties. Further details on the annotation procedure and inter-annotator agreement are provided in Appendix~\ref{annotation}.

\begin{figure}[t!]
\centering
\includegraphics[width=0.8\linewidth]{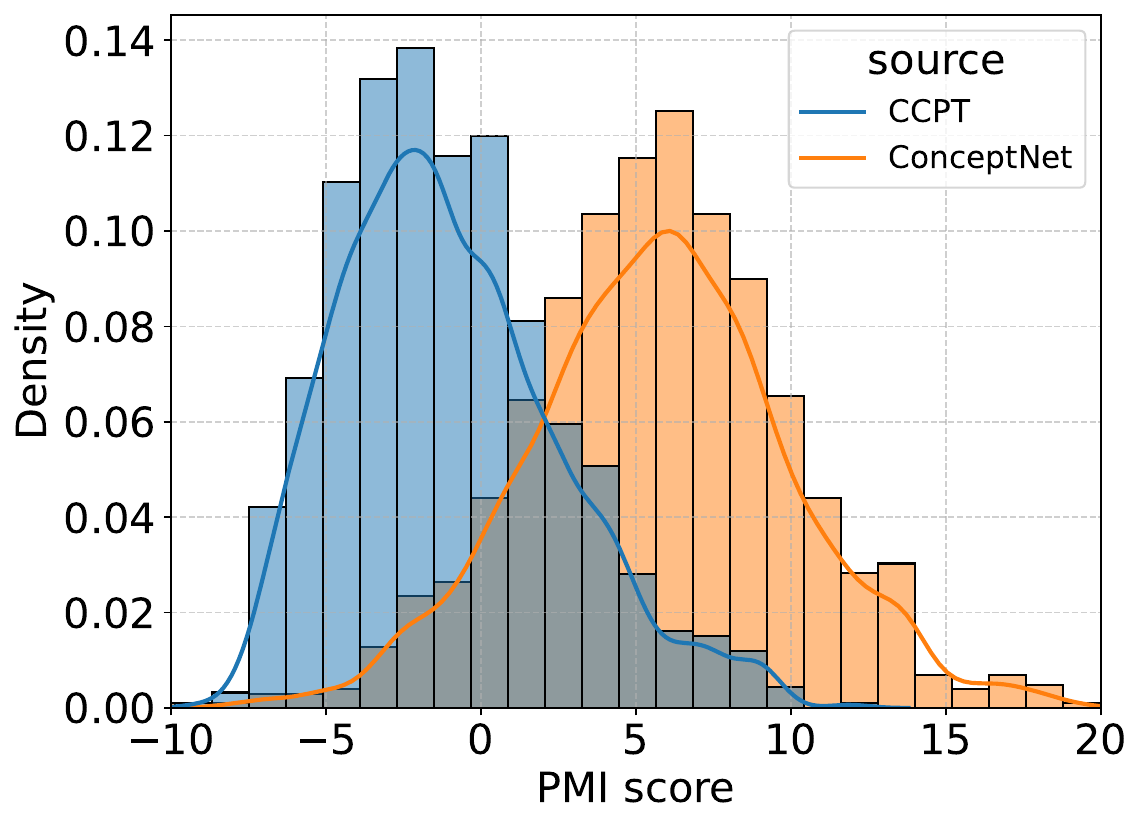}
\vspace{-1mm}
\caption{Distributions of Pointwise Mutual Information (PMI) on log-2 scale based on the Google Books N-gram Corpus.}
\label{fig:cooccurence}
\vspace{-5mm}
\end{figure}
\subsection{Co-occurence of Concepts}
\label{cooccurence_analysis}
To analyze how novel the noun phrases in our dataset is, we form a co-occurrence matrix based on PMI scores from the Google Books Ngram Corpus.
The PMI formula is:
\[
\text{PMI}(w, c) = \log_2 \frac{P(w, c)}{P(w)P(c)},
\]
where noun phrases with zero frequency are discarded. 

When comparing our dataset to bi-gram concepts from ConceptNet, Figure~\ref{fig:cooccurence} shows that combinations in our dataset have lower co-occurrence than those in ConceptNet. The average PMI for our dataset is -1.03, compared to 5.78 for ConceptNet, suggesting that combinations are more novel.

\section{Benchmarking LLMs and Humans}

We comprehensively assess 6 current LLMs with different architectures and sizes, including both open-source and closed-source models: LLaMa-3.1-70B-Instruct~\cite{dubey2024llama}, Qwen2.5-72B-Instruct~\cite{qwen2.5}, GPT-4o-20240513~\cite{openai2024gpt4o}, Claude-3.5-Sonnet~\cite{anthropic2024claude}, o1-mini-2024-09-12~\cite{openai2024o1mini}, and o1-2024-12-17~\cite{openai2024o1}. The test instances are a randomly sampled representative sample of data instances; N=200 for \textsc{Property Induction-Emergent}, N=167 for \textsc{Property Induction-Canceled}, N=167 for \textsc{Noun Phrase Completion-Emergent}.

\subsection{Methods}
To provide background knowledge on conceptual combinations, we use a consistent system prompt across all baselines and tasks. This prompt includes explanations of key concepts such as conceptual combinations, head nouns, modifiers, and different types of properties. Implementation details about model are explained in Appendix \ref{implementation}. Prompts are detailed in Appendix~\ref{prompting_details}.

\noindent\textbf{Base.} Base prompting method is to evaluate LLMs' ability to generate their responses without reasoning steps.

\noindent\textbf{Chain-of-Thought (CoT).}
We adopt chain-of-thought~\cite{kojima2022large} method to induce LLMs to generate the reasoning steps before producing the final answer by adding the phrase ``Let’s think step by step'' at the end of the questions.

\noindent\textbf{Spreading Activation (S.A.) (Ours)} Spreading activation is a cognitive model to search through networks of related ideas or concepts~\cite{anderson1983spreading, collins1975spreading}. One concept in the mind activates another concept through the pathway of association. It becomes easier for people to combine these related concepts together and understand them as a relationship.

\SetKwInput{KwInput}{Input}
\SetKwInput{KwParam}{Parameter}
\SetKwFunction{KwFunc}{Function}
\begin{algorithm}[t!]
\small 
\DontPrintSemicolon
\SetAlgoLined
\SetNoFillComment
\caption{\textsc{Spreading Activation}}\label{alg:spreading_activation}

\KwInput{
    Model \(\mathcal{M}\), 
    Initial set of seed concepts \( C_0 \), 
    Prompts \(\{p_{act}, p_{fil}\}\), 
    Concept graph \(G\)
}

\KwParam{
    Maximum iterations \( T \), 
    Convergence threshold \( \epsilon \), 
    Use language model \( U_L \), 
    Use concept graph \( U_C \)
}
\KwResult{Related concept set \( C_T \)}

\SetKwFunction{Activate}{Activate}
\SetKwFunction{Filter}{Filter}
\vspace{1em}
\# Retrieve relevant concepts \( A_c \)

\KwFunc{\Activate(\( c \))}:
    \Begin{
        \( A_c \gets \emptyset \)

        \If {\(U_{L}\)} {
            \( A_c \gets A_c \cup \mathcal{M}(p_{act}, c) \)
        }
        \If {\(U_{C}\)} {
            \( A_c \gets A_c \cup G.\text{query}(c) \)
        }     
        \Return \( A_c \)
    }
\vspace{1em}
\# Filter each concept in \( C_t \) w.r.t. \(C_0\)

\KwFunc{\Filter(\( C_t, C_0 \))}:
    \Begin{
        \Return \(\mathcal{M}(p_{fil}, C_t, C_0) \)
    }
\vspace{1em}
\For {iteration \( t = 0 \) \textbf{to} \( T \)} { 
    \( C_{t+1} \gets C_t \)
    
    \For {each concept \( c \in C_t \cup C_0 \)} {
        \( A_c \gets \Activate(c) \)  
        
        \( C_{t+1} \gets C_{t+1} \cup A_c \)
    }
    
    \(C_{t+1} \gets \Filter(C_{t+1}, C_0) \setminus C_0 \)
    
    $\displaystyle \Delta = 1 - \frac{|C_t \cap C_{t+1}|}{|C_t \cup C_{t+1}|}$
    
    \If { \( \Delta < \epsilon \) } {
        \textbf{break}
    }
}
\( C_T \gets C_{t+1} \)

\Return \( C_T \)\

\end{algorithm}

Motivated by spreading activation theory, we propose a novel method to iteratively explore relationships between concepts. In Algorithm \ref{alg:spreading_activation}, the initial concept set (\( C_0 \)) and objective (\( O \)) are first defined. For example, generating the emergent property of a peeled apple is represented as \( C_0 = \{\text{peeled, apple}\} \), with \( O \) defined as ``find relationships between ‘peeled’ and ‘apple’.'' Relevant concepts (\( A_c \)) are then activated for each component in the concept set (\( C_t \)) using either LLM or a graph-based approach such as ConceptNet. A filtering step selects components from \( C_t \) based on their relatedness to \( C_0 \). If there is no significant difference between \( C_t \) and \( C_{t+1} \), the loop terminates. After iteratively expanding the set, the LLM generates the final answer based on the intermediate concept set \( C_T \). Specifically, we set the maximum iteration steps (\( T \)) to 5 and the convergence threshold (\( \epsilon \)) to 0.1.

\noindent\textbf{Multi-Oracle.} The best result among the multiple efforts can be seen as the upper-bound performance for each LLM. For the research purpose, we include this score by selecting the best score among the multiple solutions' (N=5) scores, presented with a gray background in Table \ref{table:gen_result}.

\noindent\textbf{Gold.}
We provide the score assigned to our annotated dataset as an upper-bound performance score for our tasks.

\subsection{Human Responses}

We recruited 5 native English speaker students through offline advertisement on a university campus. None of the students knew the researchers or had heard about conceptual combinations before. We verbally introduced the meaning of conceptual combination, and by e-mail, test sheets for each task were sent. The students solved the tasks in their own time and place. 1 to 3 students solved a single test sheet. We report the best result. Question format is provided in Appendix~\ref{subsec:collection_human_solutions}.

\begin{table*}[ht!]
\centering
\resizebox{\linewidth}{!}{
\begin{tabular}{lccccccccc}
\hline
&\multicolumn{6}{c}{\textbf{\textsc{Property Induction}}}&\multicolumn{3}{c}{\textbf{\textsc{Noun Phrase Completion}}}\\\cline{2-10}
&\multicolumn{3}{c}{\textbf{(1) \textsc{Emergent}}}&\multicolumn{3}{c}{\textbf{(2) \textsc{Canceled}}}&\multicolumn{3}{c}{\textbf{(3) \textsc{Emergent}}}\\\cline{2-10}
&$R_{\mathcal{H},\mathcal{M},\mathcal{P}}$$\downarrow$&$R_{\mathcal{N},\mathcal{P}}$$\uparrow$&$\mathcal{E}$$\uparrow$&$R_{\mathcal{H},\mathcal{M},\mathcal{P}}$$\uparrow$&$R_{\mathcal{N},\mathcal{P}}$$\downarrow$&$\mathcal{C}$$\uparrow$&$R_{\mathcal{H},\mathcal{M},\mathcal{P}}$$\downarrow$&$R_{\mathcal{N},\mathcal{P}}$$\uparrow$&$\mathcal{E}$$\uparrow$\\\hline
\textit{\textbf{LLaMa3.1-70B}}&&&&&&&&&\\
Base&42.2 $\pm$ 1.8&81.4 $\pm$ 0.9&41.6 $\pm$ 0.6&64.1 $\pm$ 1.0&15.8 $\pm$ 1.2&50.3 $\pm$ 1.1&57.0 $\pm$ 1.3&\textbf{74.7 $\pm$ 0.7}&21.1 $\pm$ 1.9\\
CoT&\textul{41.8 $\pm$ 1.5}&81.4 $\pm$ 0.7&42.5 $\pm$ 1.3&61.5 $\pm$ 0.5&14.0 $\pm$ 0.8&49.5 $\pm$ 0.7&52.1 $\pm$ 1.4&\textul{73.8 $\pm$ 0.5}&24.8 $\pm$ 1.7\\
S.A.w/ LLM&\textbf{40.6 $\pm$ 0.3}&81.6 $\pm$ 0.5&\textul{43.3 $\pm$ 0.6}&66.2 $\pm$ 1.4&13.9 $\pm$ 1.0&55.2 $\pm$ 1.4&\textbf{41.2 $\pm$ 1.5}&64.2 $\pm$ 1.0&\textbf{26.6 $\pm$ 2.1}\\
S.A.w/ ConceptNet &42.1 $\pm$ 0.5&\textbf{82.0 $\pm$ 0.5}&42.6 $\pm$ 0.7&\textbf{68.2 $\pm$ 0.9}&\textul{11.4 $\pm$ 0.7}&\textul{57.9 $\pm$ 0.8}&\textul{42.9 $\pm$ 0.7}&66.0 $\pm$ 1.9&\textul{26.3 $\pm$ 1.3}\\
S.A.w/ Both &\textbf{40.6 $\pm$ 0.6}&\textul{81.7 $\pm$ 0.7}&\textbf{43.6 $\pm$ 0.3}&\textul{68.1 $\pm$ 1.3}&\textbf{11.3 $\pm$ 1.4}&\textbf{58.3 $\pm$ 2.0}&43.9 $\pm$ 0.6&64.0 $\pm$ 1.8&24.3 $\pm$ 0.8\\
\rowcolor{gray!20}Multi-Oracle&30.7 $\pm$ 0.7&89.4 $\pm$ 0.4&54.6 $\pm$ 0.5&76.3 $\pm$ 0.5&6.8 $\pm$ 0.3&64.1 $\pm$ 0.6&44.8 $\pm$ 0.6&84.1 $\pm$ 0.3&32.4 $\pm$ 0.5\\\hline
\textit{\textbf{Qwen2.5-72B}}&&&&&&&&&\\
Base&46.6 $\pm$ 0.7&82.4 $\pm$ 0.4&40.2 $\pm$ 0.8&62.7 $\pm$ 0.7&11.8 $\pm$ 0.9&52.7 $\pm$ 1.2&54.6 $\pm$ 0.6&\textbf{75.5 $\pm$ 1.1}&\textul{24.5 $\pm$ 0.8}\\
CoT&\textbf{44.4 $\pm$ 1.0}&79.3 $\pm$ 0.5&40.2 $\pm$ 0.7&62.0 $\pm$ 0.8&\textbf{9.9 $\pm$ 0.6}&54.1 $\pm$ 0.9&52.0 $\pm$ 0.9&\textul{75.0 $\pm$ 0.9}&\textbf{26.5 $\pm$ 0.6}\\
S.A.w/ LLM&45.1 $\pm$ 0.4&84.3 $\pm$ 0.4&\textbf{42.9 $\pm$ 0.7}&\textbf{76.9 $\pm$ 1.4}&\textul{13.0 $\pm$ 0.7}&\textbf{64.7 $\pm$ 1.2}&49.0 $\pm$ 0.8&68.6 $\pm$ 1.2&23.3 $\pm$ 0.3\\
S.A.w/ ConceptNet &\textul{44.8 $\pm$ 0.3}&\textul{84.5 $\pm$ 0.4}&\textbf{42.9 $\pm$ 0.2}&\textul{75.1 $\pm$ 1.8}&13.5 $\pm$ 0.9&\textul{62.4 $\pm$ 1.6}&\textbf{47.3 $\pm$ 1.6}&66.1 $\pm$ 1.8&22.4 $\pm$ 0.1\\
S.A.w/ Both &45.8 $\pm$ 0.8&\textbf{84.6 $\pm$ 0.5}&\textul{42.4 $\pm$ 1.1}&67.6 $\pm$ 1.0&\textul{13.0 $\pm$ 0.7}&56.3 $\pm$ 0.2&\textul{48.1 $\pm$ 0.3}&69.0 $\pm$ 1.3&24.2 $\pm$ 1.2\\
\rowcolor{gray!20}Multi-Oracle&40.4 $\pm$ 0.8&87.4 $\pm$ 0.3&47.5 $\pm$ 0.9&71.5 $\pm$ 0.4&6.4 $\pm$ 0.2&63.4 $\pm$ 0.5&48.3 $\pm$ 0.2&82.4 $\pm$ 0.4&31.7 $\pm$ 0.7\\\hline
\textit{\textbf{GPT-4o}}&&&&&&&&&\\
Base   &44.1 $\pm$ 0.6&83.3 $\pm$ 0.4&40.8 $\pm$ 0.7&67.5 $\pm$ 1.0&13.0 $\pm$ 0.7&55.5 $\pm$ 1.1&53.1 $\pm$ 2.0&69.8 $\pm$ 1.6&20.4 $\pm$ 1.5\\
CoT     &43.0 $\pm$ 0.5&83.1 $\pm$ 0.7&42.4 $\pm$ 0.8&68.6 $\pm$ 0.6&\textbf{11.8 $\pm$ 1.4}&57.6 $\pm$ 0.7&52.0 $\pm$ 0.4&\textbf{72.1 $\pm$ 0.2}&23.1 $\pm$ 0.4\\
S.A.w/ LLM&\textul{42.9 $\pm$ 1.6}&84.6 $\pm$ 0.7&43.8 $\pm$ 2.0&\textul{75.5 $\pm$ 1.2}&15.1 $\pm$ 1.8&60.8 $\pm$ 1.5&\textul{46.8 $\pm$ 1.0}&69.9 $\pm$ 0.3&25.5 $\pm$ 0.8\\
S.A.w/ ConceptNet &\textbf{41.5 $\pm$ 1.4}&\textul{85.8 $\pm$ 0.4}&\textbf{45.8} $\pm$ 1.7&\textbf{75.7 $\pm$ 1.2}&\textul{12.8 $\pm$ 0.6}&\textbf{63.3 $\pm$ 1.6}&\textbf{46.3 $\pm$ 0.6}&\textul{70.8 $\pm$ 1.5}&\textbf{26.9 $\pm$ 0.5}\\
S.A.w/ Both &43.0 $\pm$ 0.8&\textbf{86.6 $\pm$ 0.2}&\textul{45.2 $\pm$ 0.9}&74.0 $\pm$ 0.4&13.0 $\pm$ 1.3&\textul{62.0 $\pm$ 1.3}&47.0 $\pm$ 1.3&\textul{70.8 $\pm$ 1.5}&\textul{26.8 $\pm$ 2.0}\\
\rowcolor{gray!20}Multi-Oracle&28.9 $\pm$ 0.4&92.0 $\pm$ 0.4&57.8 $\pm$ 0.2&82.3 $\pm$ 0.6&4.5 $\pm$ 0.3&72.4 $\pm$ 0.9&35.7 $\pm$ 0.9&85.5 $\pm$ 0.7&38.9 $\pm$ 0.5\\\hline
\textit{\textbf{Sonnet-3.5}}&&&&&&&&&\\
Base&37.9 $\pm$ 0.7&78.9 $\pm$ 0.5&43.6 $\pm$ 0.9&71.1 $\pm$ 1.3&11.8 $\pm$ 0.7&59.8 $\pm$ 1.0&46.6 $\pm$ 0.5&67.8 $\pm$ 1.2&24.2 $\pm$ 1.0\\
\rowcolor{gray!20}Multi-Oracle&33.0 $\pm$ 0.8&81.2 $\pm$ 0.5&39.4 $\pm$ 0.6&84.6 $\pm$ 0.7&4.4 $\pm$ 0.1&75.2 $\pm$ 0.3&33.0 $\pm$ 0.8&81.2 $\pm$ 0.5&39.4 $\pm$ 0.6\\\hline
\textit{\textbf{o1-mini}}&&&&&&&&&\\
Base&42.1 $\pm$ 0.2&81.9 $\pm$ 0.5&42.1 $\pm$ 0.1&78.2 $\pm$ 0.1&17.8 $\pm$ 2.6&61.3 $\pm$ 2.1&47.7 $\pm$ 0.8&68.6 $\pm$ 1.2&24.5 $\pm$ 1.3\\
\rowcolor{gray!20}Multi-Oracle&26.0 $\pm$ 0.4&91.8 $\pm$ 0.2&60.3 $\pm$ 0.2&90.7 $\pm$ 0.9&5.3 $\pm$ 0.9&76.8 $\pm$ 0.9&31.3 $\pm$ 0.8&84.0 $\pm$ 0.2&44.0 $\pm$ 0.3\\\hline
\textit{\textbf{o1}}&&&&&&&&&\\
Base&37.3 $\pm$ 1.2&79.9 $\pm$ 0.9&43.5 $\pm$ 0.4&76.2 $\pm$ 1.9&7.9 $\pm$ 0.5&68.4 $\pm$ 0.9&43.8 $\pm$ 0.7&74.0 $\pm$ 1.7&32.6 $\pm$ 1.6\\
\rowcolor{gray!20}Multi-Oracle&21.9 $\pm$ 0.7&89.0 $\pm$ 0.2&60.9 $\pm$ 0.9&84.8 $\pm$ 0.9&2.3 $\pm$ 0.9&78.9 $\pm$ 0.9&30.7 $\pm$ 1.0&85.7 $\pm$ 0.7&49.7 $\pm$ 1.5\\\hline
\textit{\textbf{Human}}& 37.7 & 85.0 & 49.2 & 69.5 & 11.1 & 58.6 & 36.4 & 72.5 & 38.0\\\hline
\textit{\textbf{Gold}}&29.2&87.4&58.4&83.2&14.2&69.5&27.5&87.2&59.9\\\hline
\end{tabular}
}
\caption{Generative results on test instances, reporting average scores with their standard error of the mean (SEM). In the \textsc{Emergent Property} scenario, better emergence corresponds to lower $R_{\mathcal{H}, \mathcal{M}, \mathcal{P}} \downarrow$ and higher $R_{\mathcal{N}, \mathcal{P}} \uparrow$. In the \textsc{Canceled Property} scenario, better cancellation corresponds to higher $R_{\mathcal{H}, \mathcal{M}, \mathcal{P}} \uparrow$ and lower $R_{\mathcal{N}, \mathcal{P}} \downarrow$. \textbf{Bold score} indicates the best score; \textul{underlined score} is the second-best. Multi-Oracle in gray background represents upper-bound performance for each LLM and is not included in the rankings.}
\vspace{-1mm}
\label{table:gen_result}
\end{table*}
\begin{table*}[t!]
\resizebox{\linewidth}{!}{
\begin{tabular}{lccccccccc}
\Xhline{1.2pt} 
&\multicolumn{6}{c}{\textbf{\textsc{Property Induction}}}&\multicolumn{3}{c}{\textbf{\textsc{Noun Phrase Completion}}}\\\cline{2-10}
&\multicolumn{3}{c}{\textbf{(1) \textsc{Emergent}}}&\multicolumn{3}{c}{\textbf{(2) \textsc{Canceled}}}&\multicolumn{3}{c}{\textbf{(3) \textsc{Emergent}}}\\\cline{2-10}
\textbf{\textit{GPT-4o}}&$R_{\mathcal{H},\mathcal{M},\mathcal{P}}$$\downarrow$&$R_{\mathcal{N},\mathcal{P}}$$\uparrow$&$\mathcal{E}$$\uparrow$&$R_{\mathcal{H},\mathcal{M},\mathcal{P}}$$\uparrow$&$R_{\mathcal{N},\mathcal{P}}$$\downarrow$&$\mathcal{C}$$\uparrow$&$R_{\mathcal{H},\mathcal{M},\mathcal{P}}$$\downarrow$&$R_{\mathcal{N},\mathcal{P}}$$\uparrow$&$\mathcal{E}$$\uparrow$\\\hline
Base   & 49.2 & 84.1 & 36.3 & 70.7 & 29.4 & 43.9 & 54.3 & 69.0 & 20.4 \\
CoT     & 45.4 & 84.1 & 41.3 & 68.0 & 21.5 & 47.0 & 53.0 & \textbf{77.8} & \textul{26.7} \\
S.A.w/ LLM & \textbf{43.0} & 84.4 & \textbf{43.3} & \textbf{79.0} & 23.1 & \textbf{57.6} & 52.6 & 69.6 & 22.0 \\
S.A.w/ ConceptNet & \textul{44.8} & \textbf{86.1} & \textul{42.2} & 75.9 & \textbf{19.6} & \textul{56.5} & \textbf{48.7} & 65.4 & 20.2 \\
S.A.w/ Both & 47.4 & \textul{84.6} & 40.2 & \textul{77.8} & 22.8 & 56.1 & \textul{50.9} & \textul{75.9} & \textbf{29.6} \\\hline
\textbf{\textit{Human}}& 42.8 & 84.8 & 44.3 & 73.5 & 22.4 & 52.6 & 43.7 & 79.8 & 39.0 \\\hline
\textbf{\textit{Gold}} & 34.4 & 87.2 & 52.8 & 79.6 & 21.1 & 59.0 & 33.0 & 90.6 & 57.6 \\\Xhline{1.2pt} 
\end{tabular}
}
\caption{Human evaluation of generative results on sub-sampled test instances (N=30). The relevance scores are calculated by the average scores of two annotators. \textbf{Bold numbers} indicate the best scores, while \underline{underline} numbers are the second-best.}
\label{table:gen_result_human}
\vspace{-5mm}
\end{table*}

\subsection{Evaluation Metric}\label{sec:eval_metric}
\noindent\textbf{Generative Tasks.}
It can be challenging to determine whether emergence or cancellation occurs. For instance, an overripe apple may be more strongly associated with red than a regular apple, but this doesn't clearly indicate emergence. Similarly, determining cancellation presents the same difficulty. To address this, we propose two metrics based on a continuous relevance scoring. Both human judges and the LLM-as-a-judge approach~\cite{zheng2023judging} are utilized to evaluate open-ended generative responses. These metrics are applied to the \textsc{Property Induction} and \textsc{Noun Phrase Completion}.
\setlength{\itemsep}{0pt} \begin{itemize}
    \item The \textbf{emergence score} ($\mathcal{E}$) measures how suddenly properties arise when concepts are combined, compared to the properties of the individual concepts.
    \vspace{-1mm}
    \item The \textbf{cancellation score} ($\mathcal{C}$) reflects how much a property is diminished or canceled when concepts are combined, compared to their individual properties.
\end{itemize}

Specifically, each score is defined as
\begin{align}
&R_{\mathcal{H},\mathcal{M},\mathcal{P}} = \max \{R_{\mathcal{H}, \mathcal{P}}, R_{\mathcal{M}, \mathcal{P}}\}, \\
    &\mathcal{E} = \max \{R_{\mathcal{N}, \mathcal{P}}-R_{\mathcal{H},\mathcal{M},\mathcal{P}}, 0\}, \\
    &\mathcal{C} = \max \{R_{\mathcal{H},\mathcal{M},\mathcal{P}} - R_{\mathcal{N}, \mathcal{P}}, 0\},
\end{align}
Here, \( R_{\mathcal{X},\mathcal{P}} \) represents the relevance score, indicating how strongly concept \( \mathcal{X} \) possesses property \( \mathcal{P} \), measured by human judges and GPT-4o~\cite{openai2024gpt4o} on a scale from 0 to 1. Exceptionally, for the \textsc{Noun Phrase Completion}, cancellation score is not measured since it easily occurs by adding antonym of given canceled property (e.g., to cancel ``yellowness'' of ``banana'', add ``brown'' to ``banana''). Detailed instructions for both manual evaluation and LLM-as-a-judge are provided in Table \ref{tab:prompt_judge} and Figure \ref{fig:mturk_relevance_tagging_instruction}.

\noindent\textbf{Classification Task.}
In this classification setup, we evaluate performance based on accuracy (\%) in the \textsc{Property Type Prediction}. The classification involves four categories: ``canceled property'', ``emergent property'', ``component property'', and ``others property''. The ``others property'' category is specifically for properties unrelated to the combination and its components.

\section{Results}

\subsection{Generative Task Result}

Table~\ref{table:gen_result} presents the experimental results in the generative setting for the two tasks, \textsc{Property Induction} and \textsc{Noun Phrase Completion}, evaluated based on LLM-as-a-judge. Table~\ref{table:gen_result_human} shows the corresponding results based on human-judge evaluations for the same tasks.

\noindent\textbf{Which property do the LLMs generate better: emergent or canceled?} As shown in Table \ref{table:gen_result}-(1) and (2), all baseline models find it more challenging to generate \textit{emergent properties} than \textit{canceled properties}. Humans outperform LLMs in generating emergent properties from noun phrases but perform worse in handling canceled properties. For emergent properties, GPT-4o-S.A. w/ ConceptNet achieves a significantly higher relatedness score between the property and each component, $\mathcal{R}_{\mathcal{H}, \mathcal{M}, \mathcal{P}}$, exceeding the gold score by 12.3 points. In contrast, the relatedness score between the property and the noun phrase, $\mathcal{R}_{\mathcal{N}, \mathcal{P}}$, is only 1.6 points lower. While LLMs generate properties that align well with the noun phrase, they tend to rely on properties already associated with the individual components.

\noindent\textbf{How well do LLMs create conceptual combinations?} In Table \ref{table:gen_result}-(3), it is difficult for all baselines to come up with a modifier $\mathcal{M}$ that lacks a given property on its own but exhibits that property when combined with a head noun. The emergence scores $\mathcal{E}$ for this task are 27 or lower across all models. Humans are better than LLMs at generating combinations that exhibit a given emergent property.

\noindent\textbf{Which generative task do LLMs excel at?} We compared two tasks: generating emergent properties (Table \ref{table:gen_result}-(1)) and generating a noun phrase (Table \ref{table:gen_result}-(1)). All baseline models achieve lower emergence scores (\(\mathcal{E}\)) in the noun phrase completion task. This suggests that LLMs find it more challenging to create a noun phrase that accurately captures an emergent property than to identify an emergent property from a noun phrase.

\subsection{Classification Task Result}\label{subsec:classification_task_result}

\begin{table}[!t]
    \centering
    \small
    \renewcommand{\arraystretch}{1.5}
    \setlength{\aboverulesep}{0pt}
    \setlength{\belowrulesep}{0pt}
    \setlength{\arrayrulewidth}{0.5pt}  
    \resizebox{\linewidth}{!}{
    \begin{tabular}{|cc|c|c|c|c|} 
    \hline
    \multirow{2}{*}{} & \multirow{2}{*}{} & \multicolumn{4}{c|}{\textbf{Predicted Types}} \\ \cline{3-6} 
     &  & \textbf{Emergent} & \textbf{Component} & \textbf{Canceled} & \textbf{Others} \\ \hline
    \multicolumn{1}{|c|}{\multirow{4}{*}{\rotatebox[origin=c]{90}{\textbf{Actual Types}}}} 
    & \textbf{Emergent} & \cellcolor{green!25}\textbf{90.0} & \cellcolor{green!25} 4.4 & 2.0 & 3.6 \\ \hhline{|~|-|-|-|-|-|}
    \multicolumn{1}{|c|}{} 
    & \textbf{Component} & \cellcolor{green!25}59.6 & \cellcolor{green!25}\textbf{37.2} & 1.2 & 2.0 \\ \hhline{|~|-|-|-|-|-|}
    \multicolumn{1}{|c|}{} 
    & \textbf{Canceled} & 13.6 & 15.6 & \cellcolor{red!25}\textbf{45.2} & \cellcolor{red!25}25.6 \\ \hhline{|~|-|-|-|-|-|}
    \multicolumn{1}{|c|}{} 
    & \textbf{Others} & 26.0 & 5.6 & \cellcolor{red!25}15.2 & \cellcolor{red!25}\textbf{53.2} \\ \hline
    \end{tabular}}
    \caption{Classification results of GPT-4o for property type prediction based on 1000 sampled instances (250 samples per property type). We present the average accuracy (\%). Cells with a green background indicate cases where the model correctly predicts that combinations possess a property (95.6\% acc), while cells with a red background indicate cases where it correctly predicts that combinations do not possess a property (69.6\% acc).}
    \label{tab:property_type_prediction}
    \vspace{-5mm}
\end{table}

Table \ref{tab:property_type_prediction} presents the experimental results for the \textsc{Property Type Prediction}. Given a noun phrase and a property, GPT-4o classifies the property into one of four categories: emergent, component, canceled, or others (where others is unrelated to both the combination and its components).

\noindent\textbf{Do LLMs identify the type of property well?}
In determining whether a noun phrase has a given property, GPT-4o achieves an accuracy of 82.6\% \(\left((95.6\% + 69.6\%) \div 2\right)\). However, in predicting the type of property, GPT-4o is correct only 56.4\% of the time, falling behind human accuracy (81\%), as noted in Appendix \ref{subsec:property_type_mturk_detail}. Compared to its accuracy in identifying emergent properties, its performance across other property types lags significantly. These results suggest room for improvement in understanding different property types.

\section{Analysis}

\subsection{Relevance between LLM-as-a-judge metric and Manual Evaluation}
\begin{figure}[t!]
\centering
\includegraphics[width=0.8\linewidth]{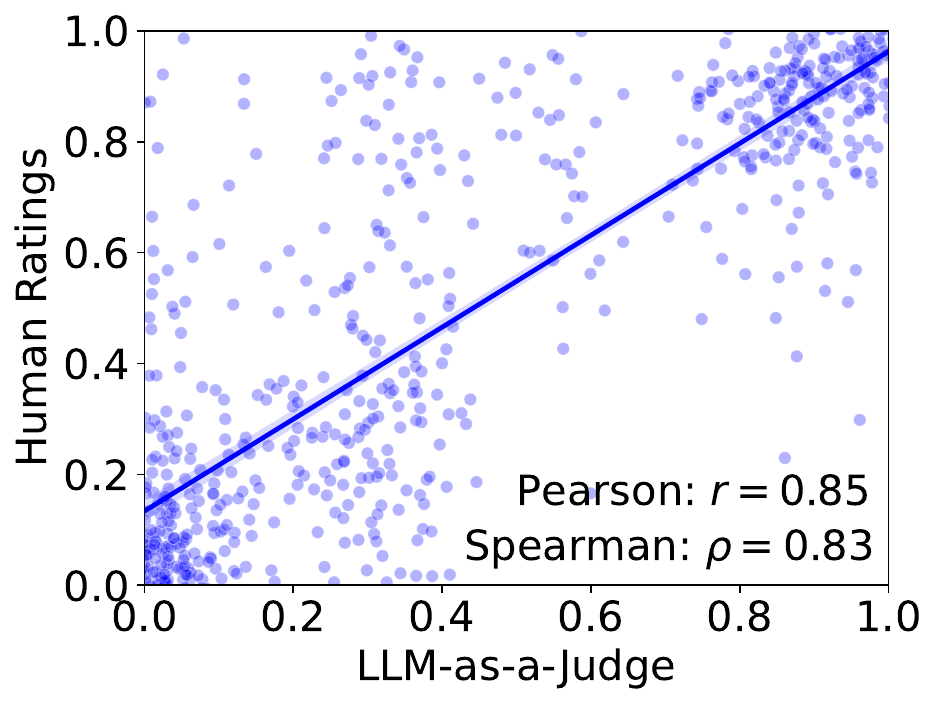}
\vspace{-1mm}
\caption{Correlation between LLM-as-a-judge and human ratings in relevance score, assessing how strong a concept \( \mathcal{X} \) possesses a property \( \mathcal{P} \). To avoid overlapping points, random jitters sampled from $\mathcal{N}(0, 0.05^2)$ are added to LLM-as-a-judge and human ratings after fitting the regression.}
\label{fig:human-llm-relation}
\vspace{-5mm}
\end{figure}
\label{subsec:relevance_llm_human}
As shown in Figure \ref{fig:human-llm-relation}, we compare the LLM judge's metrics with human evaluations to verify the agreement between them. We randomly selected 300 pairs from \mydata, which consists of pairs from $\mathcal{H}-\mathcal{P}$, $\mathcal{M}-\mathcal{P}$, and $\mathcal{N}-\mathcal{P}$, covering both emergent properties (50 samples) and canceled properties (50 samples). Human raters, recruited through Amazon Mechanical Turk, as detailed in Section~\ref{subsec:evaluation_between_llm_as_a_judge_and_human_judge}, are asked to rate the relevance of each pair using the same instructions provided to the LLM judge. Each problem is rated by three different raters.
We calculated the Pearson and Spearman correlation coefficients between the LLM judge's scores and human ratings, which are 0.85 and 0.83, respectively. These strong correlations indicate a high level of agreement, demonstrating the effectiveness of using LLM for relevance scoring.

\subsection{Analysis for Spread Activation Method}

First, to evaluate the impact of multiple iterations on performance, we adjust \( T \) from 1 to 5 and gather the answers for each \( C_T \). Table \ref{iteration_sa_performance} highlights that as the number of iterations increases, the performance of each task gradually improves. This indicates that iteratively propagating relevant concepts is effective in our conceptual combination tasks.

Second, our spread activation method consists of the functions \(\Activate(\cdot)\) and \(\Filter(\cdot, \cdot)\). To investigate the contribution of \(\Filter\) on performance, we conducted an ablation study by comparing scores with and without \(\Filter\) in the spread activation method. As shown in Table \ref{iteration_sa_filter}, performance decreases when \(\Filter\) is removed. This suggests that iteratively eliminating distractors improves overall performance.




\section{Conclusion}
In conclusion, our work introduces \mydata, a conceptual combination dataset designed to evaluate LLMs' ability to process conceptual combinations. \mydata~comprises 12,315 annotated instances of noun phrases, properties, and property types. Based on \mydata, we propose three downstream tasks: property induction, noun phrase completion, and property type prediction. To assess generative performance, we introduce two automatic evaluation metrics—emergence and cancellation scoring—which closely align with human evaluations. Additionally, we propose a novel evaluation method inspired by cognitive psychology models. Our findings indicate that (1) LLMs struggle more with generating emergent properties than with canceled ones. Furthermore, (2) generating a noun phrase that exhibits a given emergent property proves more challenging than generating the property itself. Notably, (3) GPT-4o struggles to determine property types compared to humans. Finally, (4) our proposed spread activation method achieves the highest performance among the evaluated approaches due to its iterative retrieval of relevant concepts and filtering process.
\begin{table}[t!]
    \centering
    \begin{tabular}{lccccc}
        \toprule
        \textbf{Task} & $C_1$ & $C_2$ & $C_3$ & $C_4$ & $C_5$ \\
        \midrule
        PI-Eme. & 45.1 & 45.4 & 45.2 & \textbf{45.9} & 45.8 \\
        PI-Can. & 60.5 & 61.1 & 61.1 & 62.7 & \textbf{63.3} \\
        NPC-Eme. & 26.0 & 26.2 & \textbf{26.9} & 26.7 & \textbf{26.9} \\
        \bottomrule
    \end{tabular}
    \caption{Iteration-wise score improvement of GPT-4o + S.A.w/ConceptNet. PI and NPC denote the \textsc{Property Induction} and \textsc{Noun Phrase Completion}, respectively. \( C_t \) represents the utilization of \( C_t \) to generate the final answer. Each score in the table corresponds to either \( \mathcal{E} \) or \( \mathcal{C} \), depending on the property type.}
    \label{iteration_sa_filter}
\end{table}
\begin{table}[t!]
    \centering
    \begin{tabular}{lccc}
        \toprule
        \textbf{Task} & w/o $\Filter$ & w/ $\Filter$ & $\Delta$ \\
        \midrule
        PI-Eme. & 44.6 $\pm$ 0.9 &\textbf{45.8 $\pm$ 1.7} & $\uparrow$ 1.2\\
        PI-Can. & 55.6 $\pm$ 1.0 &\textbf{63.3 $\pm$ 1.6}  & $\uparrow$ 7.7\\
        NPC-Eme. & 26.0 $\pm$ 1.6 & \textbf{26.9 $\pm$ 0.5}  & $\uparrow$ 0.9\\
        \bottomrule
    \end{tabular}
    \caption{Impact of filtering in the spread activation method. The columns compare performance with and without filtering, highlighting the absolute score change ($\Delta$).}
    \label{iteration_sa_performance}
\end{table}

\newpage
\section*{Limitations}
We acknowledge few potential limitations of our research. (1) There is fundamental diversity in people's mental representations of the world, especially across cultural contexts. The notions of ``property'' and ``property type'' in our dataset may implicitly reflect the commonsense knowledge of the annotators' demographic group. Moreover, the approval of our data through the MTurk study may primarily reflect the commonsense of the Turkers. Future work could further explore the relationship between conceptual combination understanding and cultural divergence in concepts. (2) Homonyms can introduce misleading effects on the evaluation process. If the grader misinterprets the definition of concept from the solver's intention, the solver's performance may not be fully captured. (3) Our data generation pipeline employs comparative sentences for efficiency. However, the inherent nature of comparative sentences may introduce skewness, favoring certain types of properties over others.

\section*{Ethics Statement}
The authors checked all examples and found no personal identifying information (PII). As addressed in \S~\ref{data construction}, we also eliminated the offensive contents manually. 

\section*{Acknowledgements} 
We thank the anonymous reviewers and Yunah Jang for their valuable comments.
This work was supported by 
the Institute of Information \& Communications Technology Planning \& Evaluation (IITP) grant funded by the Korea government (MSIT) (No.~RS-2019-II191082, SW StarLab
), 
the National Research Foundation of Korea (NRF) grant funded by the Korea government (MSIT) (No.~2023R1A2C2005573), 
the Basic Science Research Program through the National Research Foundation of Korea (NRF) funded by the Ministry of Education(RS-2023-00274280), 
 Korea Radio Promotion Association (Development of Intelligent Docent Service for Information-Disadvantaged Groups), 
and the SNU-Global Excellence Research Center establishment project. 
Gunhee Kim is the corresponding author.

\bibliography{anthology, custom}

\newpage

\newpage

\appendix

\section{Data Details}
\subsection{Textual Corpora}
\label{textual_corpora}
We utilize English textual corpora from datasets and websites across different domains including books and news. Our source is comprised of Toronto Book Corpus~\cite{zhu2015aligning}, WritingPrompts~\cite{fan2018hierarchical}, Gutenberg Corpus~\cite{gerlach2020standardized}, MediaSum News Dialogue~\cite{zhu-etal-2021-mediasum}, Wikipedia-240823\footnote{https://dumps.wikimedia.org/enwiki/latest/}, r/FUNNY\footnote{https://www.reddit.com/r/funny/}, r/FantasyWriters\footnote{https://www.reddit.com/r/fantasywriters/}, and r/OCPoetry\footnote{https://www.reddit.com/r/OCPoetry/}.

\subsection{Dataset Construction Statistics}
\label{data_construction_stat}
In Table \ref{table:construction_statistics}, we describe the size of the dataset during the data collection pipeline.

\subsection{Annotation Procedure}
\label{annotation}

We hire skilled raters in Amazon Mechanical Turk (MTurk) to annotate the property type. For emergent properties and canceled properties, we assign three annotators per instance, and used majority label for the final property type. For component properties, due to cost constraints, we assign one annotator per instance.

Annotators were selected based on their success in a qualification task (Human Intelligence Task or HIT), which assessed their ability to distinguish property types. This qualification task included 10 thoroughly verified examples, with a payment of \$1.00. We required annotators to be from English-speaking countries (AU, CA, NZ, US, GB), have completed more than 10,000 HITs, and maintain a HIT approval rate greater than 98\%.

\begin{figure*}[t!]
\centering
\includegraphics[width=\linewidth]{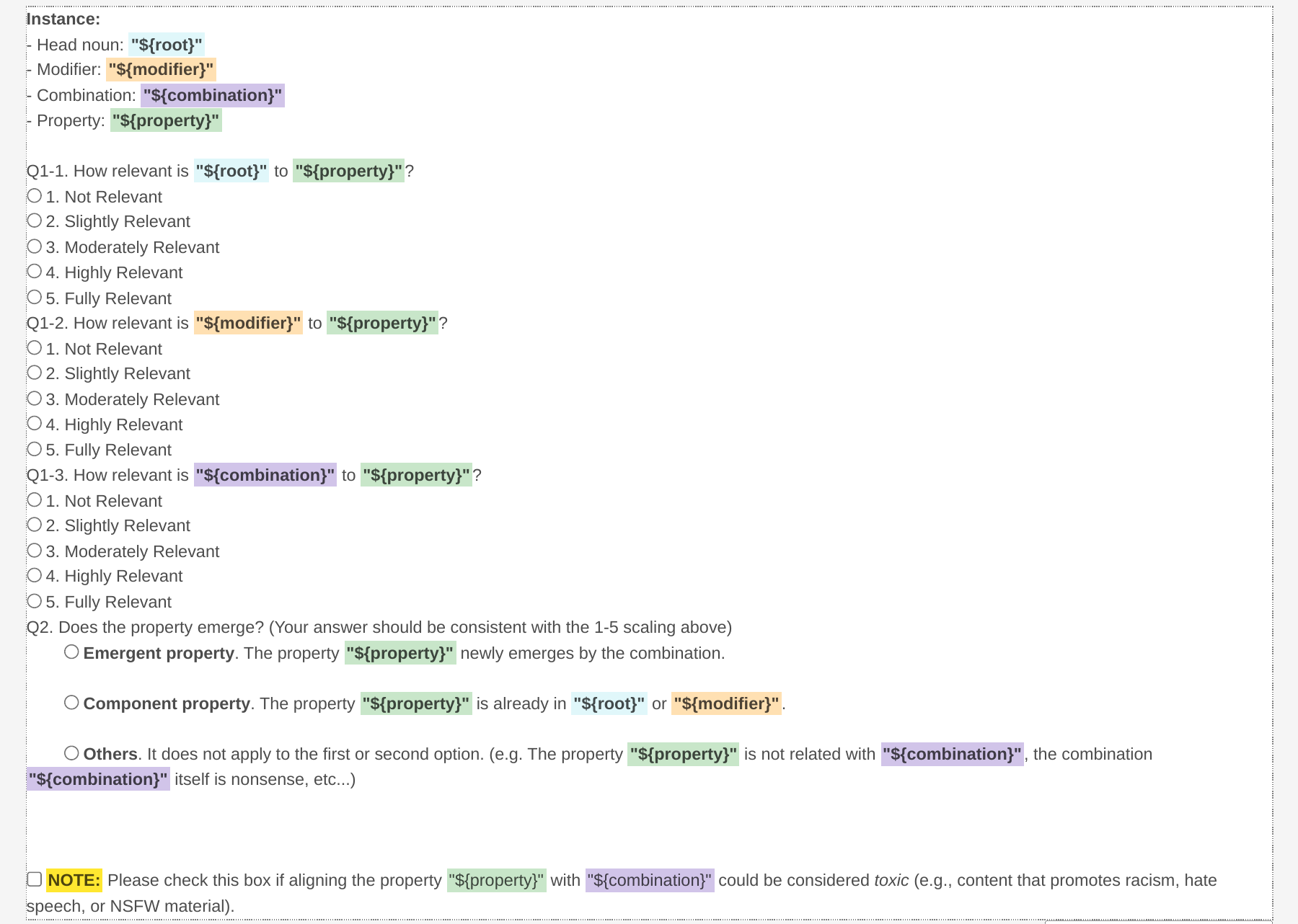}
\caption{Instructions provided for annotators of emergent property data candidates.}
\label{fig:mturk_instruction_emergent_property_candidate}
\end{figure*}

\begin{figure*}[t!]
\centering
\includegraphics[width=\linewidth]{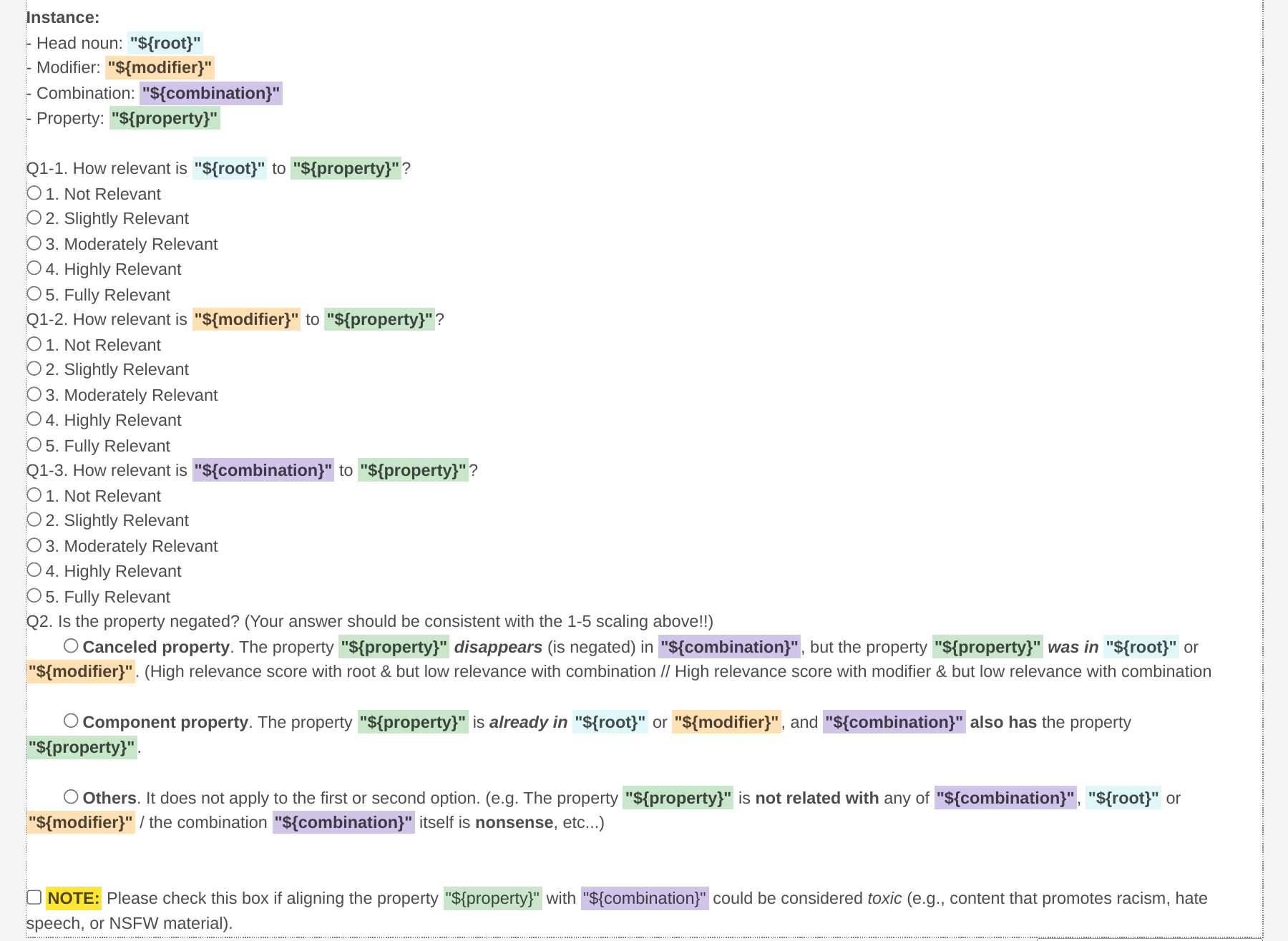}
\caption{Instructions provided for annotators of canceled property data candidates.}
\label{fig:mturk_instruction_canceled_property_candidate}
\end{figure*}

After qualification, annotators received detailed instructions on conceptual combinations with examples. They answered three questions: (1) Relevance Rating – Annotators rated the relevance of each concept ($\mathcal{N},\mathcal{H},\mathcal{M}$) to a given property ($\mathcal{P}$) to encourage careful consideration of the data. (2). Property Type Annotation – For emergent and component properties, annotators chose: (1) emergent, (2) component, or (3) other. For canceled properties, they chose: (1) canceled, (2) component, or (3) other, as illustrated in Figures~\ref{fig:mturk_instruction_emergent_property_candidate} and \ref{fig:mturk_instruction_canceled_property_candidate}. For component property candidates there is a third question about locating the source of the property from the following options: (1) noun phrase $\mathcal{N}$, (2) head noun $\mathcal{H}$ and noun phrase $\mathcal{N}$, (3) modifier $\mathcal{H}$ and noun phrase $\mathcal{N}$, (4) All of $\mathcal{N,H,M}$, (5) others. (3) Toxicity: The annotators are asked to check a box if the data is toxic. Finally, the Fleiss' $\kappa$ score of three-way classification during the annotation phase is 0.312 for emergent property data and 0.410 for canceled property data. In \mydata, we contain disaggregated human annotations for all questions.

Next, we calculate inter-annotator result for the final dataset. For emergent and component properties, we randomly selected 100 samples for this task, each reviewed by two annotators who had not participated in the original annotation. The Fleiss' $\kappa$ score for this binary classification task was 0.498, indicating agreement levels ranging from ``moderate'' to ``substantial.'' For canceled properties, we also selected 100 samples, with two annotators reviewing each example. The Fleiss' $\kappa$ score for this binary classification is 0.505, indicating agreement between ``moderate'' and ``substantial'' levels.

\section{Experimental Details}
\subsection{Implementation Details}
\label{implementation}
We collect responses using Nucleus sampling with $\mathcal{T}=0.7$ and $p=0.95$, by selecting the most likely sequence. Model responses are generated using three different seeds, and we report the average scores along with the standard error of the mean (SEM).

\begin{table}[!t]
\resizebox{\linewidth}{!}{
\begin{tabular}{lc}
\toprule
& \# of Data  \\\midrule
Corpora & \textasciitilde 51,000,000 \\
Extracting Comparative Sentences & 136,300 \\
Extracting and Filtering  & 41,660 \\
\makecell[l]{Property Type Tagging (w/ Toxic Flitering)}  & 12,315 \\\bottomrule
\end{tabular}
}
\caption{The size of the dataset during the data collection pipeline is represented by the number of remaining sentences or noun phrases at each step. The term ``\# of data'' denotes the amount of data left after each processing stage.}
\label{table:construction_statistics}
\end{table}

\subsection{Details in prompting}
\label{prompting_details}
In this section, we explain our task instruction templates. For each method, we give a correct and wrong answers for one example.

For the system prompt, refer to Table~\ref{tab:prompt_system}. For the prompt given to Base method, refer to Table~\ref{tab:npc_prompt_base} and \ref{tab:pi_prompt_base}. For the CoT-prompting, refer to Table~\ref{tab:npc_prompt_cot} and \ref{tab:pi_prompt_cot}. Spread-Activation method uses a prompt in Table~\ref{tab:npc_prompt_sa} and \ref{tab:pi_prompt_sa}. For the propery type prediction task, refer to Table~\ref{tab:pt_prompt}.

Our prompt for LLM-as-a-judge is in Table~\ref{tab:prompt_judge}. The prompt provides the fine-grained scoring criteria from 1 to 10 and three scored examples.

\begin{table*}[htbp]
\scriptsize
\centering
\begin{tabular}{@{}p{\linewidth}@{}}
\toprule
\textbf{System Prompt}\\
\midrule
Conceptual combination is a task that combines two concepts, which can result in new properties. It involves a head noun, a modifier, and corresponding properties. Here's the definition of each component:\\
\\
1. Head Noun: The original concept in the conceptual combination.\\
2. Modifier: The word that modify head noun to create a new conceptual combination.\\
3. Component Property: A property inherent to individual concepts (head noun or modifier).\\
4. Emergent Property: A new property that arises from the combination of the head noun and the modifier. This property does not exist in either concept individually (head noun or modifier) but emerge in conceptual combination.\\
5. Canceled Property: A property that is inherent to individual concept (head noun or modifier) and negated due to the combination.\\
\bottomrule
\end{tabular}
    \caption{System prompt for background of conceptual combination.}
    \label{tab:prompt_system}
\end{table*}
\begin{table*}[htbp]
\scriptsize
\centering
\begin{tabular}{@{}p{\linewidth}@{}}
\toprule
\textbf{Task Instruction for Base Method in Noun Phrase Completion}\\
\midrule
Instructions:\\
1. You are given a head noun and emergent property. Your task is to generate a conceptual combination by adding one modifier.\\
2. You can use function word without any constraint.\\
3. Modifier should not have the given emergent property on its own, but the combination exhibits the emergent property.\\
4. Use the previous examples to learn the task.\\
5. Answer in dictionary format: \{\{"combination": "\{\{generated\_combination\}\}", "modifier": "\{\{generated\_modifier\}\}"\}\}. Do not include other formatting.\\
\\
<Example 1>\\
- Head noun: apple\\
- Emergent property: unappetizing\\
- Correct answer: \{\{"combination": "brown apple", "modifier": "brown"\}\}\\
\\
Above answer is correct because each component "brown" and "apple" do not possess "unappetizing" but "brown apple" does.\\
\\
<Example 2>\\
- Head noun: banknote\\
- Emergent property: useless\\
- Wrong answer: \{\{"combination": "burned banknote", "modifier": "burned"\}\}\\
\\
Above answer is wrong because modifier "burned" directly elicit property "useless". Avoid modifier which has given property in itself.\\
\\
\\
Then let's begin:\\
\bottomrule
\end{tabular}
    \caption{Task instruction for Base Method in \textsc{Noun Phrase Completion}.}
\label{tab:npc_prompt_base}
\end{table*}
\begin{table*}[htbp]
\scriptsize
\centering
\begin{tabular}{@{}p{\linewidth}@{}}
\toprule
\textbf{Task Instruction for Chain-of-Thought Method in Noun Phrase Completion}\\
\midrule
Instructions:\\
1. You are given a head noun and emergent property. Your task is to generate a conceptual combination by adding one modifier.\\
2. You can use function word without any constraint.\\
3. Modifier should not have the given emergent property on its own, but the combination exhibits the emergent property.\\
4. Come up with your reasoning process before giving your final answer.\\
5. Use the previous examples to learn the task.\\
6. Answer in dictionary format: \{\{"combination": "\{\{generated\_combination\}\}", "modifier": "\{\{generated\_modifier\}\}"\}\}. Do not include other formatting.\\
    \\
<Example 1>\\
- Head noun: apple\\
- Emergent property: unappetizing\\
- Correct answer: Let's think step-by-step. A typical apple is fresh and appetizing, but certain modifications can make it unappetizing. Factors like discoloration, decay, or unusual texture can contribute to this perception. A brown apple, for instance, appears spoiled or oxidized, making it less appealing to eat. So the answer is \{\{"combination": "brown apple", "modifier": "brown"\}\}\\
\\
Above answer is correct because each component "brown" and "apple" do not possess "unappetizing" but "brown apple" does.\\
\\
<Example 2>\\
- Head noun: banknote\\
- Emergent property: useless\\
- Wrong answer: Let's think step-by-step. A typical banknote has value and can be used for transactions, but certain modifications can make it useless. Burning a banknote destroys its structure, making it unrecognizable and invalid as currency. So the answer is \{\{"combination": "burned banknote", "modifier": "burned"\}\}\\
\\
Above answer is wrong because modifier "burned" directly elicit property "useless". Avoid modifier which has given property in itself.\\
\\
\\
Then let's begin:\\
\bottomrule
\end{tabular}
    \caption{Task instruction for CoT Method in \textsc{Noun Phrase Completion}.}
    \label{tab:npc_prompt_cot}
\end{table*}
\begin{table*}[htbp]
\scriptsize
\centering
\begin{tabular}{@{}p{\linewidth}@{}}
\toprule
\textbf{Task Instruction for Spread Activation Method in Noun Phrase Completion}\\
\midrule
Instructions:\\
1. You are given a head noun, emergent property and a set of relevant concepts to solve a task. Your task is to generate a conceptual combination by adding one modifier.\\
2. You can use function word without any constraint.\\
3. Modifier should not have the given emergent property on its own, but the combination exhibits the emergent property.\\
4. Come up with your reasoning process before giving your final answer.\\
5. Final answer should follow dictionary format: \{\{"combination": "\{\{generated\_combination\}\}", "modifier": "\{\{generated\_modifier\}\}"\}\}. Do not use other formatting.\\
    \\
<Example 1>\\
- Head noun: apple\\
- Emergent property: unappetizing\\
- Relevant concepts: ['bland', 'bitter', 'inedible', 'insipid', 'unappealing', 'unpalatable']\\
- Correct answer: The goal is to find a modifier that does not inherently have the emergent property "unappetizing," but do when combined with "apple". Related concepts such as bitter, inedible or unpalatable make apple unappetizing. To represent bitter apple, "yellow" can be used as a modifier. But yellow is somewhat related to bitter because of the color of lemons. To represent inedible or unpalatable apple, "plastic" or "brown" can be used as a modifier. However plastic is directly related to inedible. "Brown" as a modifier doesn't imply inedible on its own, but when paired with "apple," it suggest an inedible state. So the answer is \{\{"combination": "brown apple", "modifier": "brown"\}\}\\
\\
Above answer is correct because each component "brown" and "apple" do not possess "unappetizing" but "brown apple" does.\\
\\
<Example 2>\\
- Head noun: banknote\\
- Emergent property: useless\\
- Relevant concepts: ['counterfeit', 'worthless', 'ineffective', 'futile', 'meaningless', 'pointless']\\
- Wrong answer: The goal is to find a modifier that does not inherently have the emergent property "useless," but do when combined with "banknote". Related concepts such as counterfeit or worthless make banknote useless. To represent counterfeit banknote, "fake" can be used as a modifier. But fake is somewhat related to useless because of the meaning. To represent worthless banknote, "burned" can be used as a modifier. So the answer is \{\{"combination": "burned banknote", "modifier": "burned"\}\}\\
\\
Above answer is wrong because modifier "burned" directly elicit property "useless".  Avoid modifier which elicit given property in itself.\\
\\
\\
Then let's begin:\\
\bottomrule
\end{tabular}
    \caption{Task instruction for Spread Activation Method in \textsc{Noun Phrase Completion}.}
    \label{tab:npc_prompt_sa}
\end{table*}
\begin{table*}[htbp]
\scriptsize
\centering
\begin{tabular}{@{}p{\linewidth}@{}}
\toprule
\textbf{Task Instruction for Base Method in Property Induction (Emergent)}\\
\midrule
Instructions:\\
1. You are given a combination of concepts. Your task is to generate emergent property of a combination.\\
2. Find a property that does not belong to any of the individual component in the combination but emerges when the words are combined.\\
3. Use the previous examples to learn how to do this.\\
4. Answer in dictionary format: \{\{"property": "\{\{generated\_property\}\}"\}\}. Do not include other formatting.\\
    \\
<Example 1>\\
- Combination: Brown apple\\
- Correct answer: \{\{"property": "unappetizing"\}\}\\
\\
Above answer is correct because property "unappetizing" does not belong to either "brown" and "apple", but belong to brown apple\\
\\
<Example 2>\\
- Combination: burned banknote\\
- Wrong answer: \{\{"property": "useless"\}\}\\
\\
Above answer is wrong because "burned" something can directly belong to property "useless". Ensure that the emergent property you generate does not directly describe any of the individual words but is a characteristic of the combination as a whole.\\
\\
Then let's begin:\\
\bottomrule
\end{tabular}
    \caption{Task instruction for Base Method in \textsc{Property Induction}.}
\label{tab:pi_prompt_base}
\end{table*}
\begin{table*}[htbp]
\scriptsize
\centering
\begin{tabular}{@{}p{\linewidth}@{}}
\toprule
\textbf{Task Instruction for Chain-of-Thought Method in Property Induction (Emergent)}\\
\midrule
Instructions:\\
1. You are given a combination of concepts. Your task is to generate emergent property of a combination.\\
2. Find a property that does not belong to any of the individual component in the combination but emerges when the words are combined.\\
3. Use the previous examples to learn how to do this.\\
4. Come up with your reasoning process before giving your final answer.\\
5. Answer in dictionary format: \{\{"property": "\{\{generated\_property\}\}"\}\}. Do not include other formatting.\\
    \\
<Example 1>\\
- Combination: Brown apple\\
- Correct answer: Let's think step-by-step. "Brown" signifies a color change due to aging, oxidation, or decay, indicating the apple is no longer fresh. "Apple" is a fruit that is typically appealing when fresh, but browning suggests overripeness or spoilage. Concepts like spectrum (color change) and growth (life cycle of the fruit) reinforce this transition. So the answer is \{\{"property": "unappetizing"\}\}\\
\\
Above answer is correct because property "unappetizing" does not belong to either "brown" and "apple", but belong to brown apple\\
\\
<Example 2>\\
- Combination: burned banknote\\
- Wrong answer: Let's think step-by-step. Individually, "burned" does not mean something is useless. A "banknote" by itself is valuable and serves as a medium of exchange. However, when combined, a "burned banknote" implies that the note is damaged beyond recognition, making it invalid for transactions and effectively useless as currency. So the answer is  \{\{"property": "useless"\}\}\\
\\
Above answer is wrong because "burned" something can directly belong to property "useless". Ensure that the emergent property you generate does not directly describe any of the individual words but is a characteristic of the combination as a whole.\\
\\
Then let's begin:\\
\bottomrule
\end{tabular}
    \caption{Task instruction for CoT Method in \textsc{Property Induction}.}
    \label{tab:pi_prompt_cot}
\end{table*}
\begin{table*}[htbp]
\scriptsize
\centering
\begin{tabular}{@{}p{\linewidth}@{}}
\toprule
\textbf{Task Instruction for Spread Activation Method in Noun Phrase Completion (Emergent)}\\
\midrule
Instructions:\\
1. You are given a combination of concepts and a set of relevant concepts to solve a task. Your task is to generate emergent property of a combination.\\
2. Find a property that does not belong to any of the individual component in the combination but emerges when the words are combined.\\
3. Come up with your reasoning process before giving your final answer.\\
4. Final answer should follow dictionary format: \{\{"property": "\{\{generated\_property\}\}"\}\}. Do not include other formatting.\\
    \\
<Example 1>\\
- Combination: Brown apple\\
- Relevant concepts: ['fruit', 'apple', 'core', 'cider']\\
- Correct answer: The goal is to find an emergent property of "brown apple" that does not exist in "brown" or "apple" individually. "Fruit" and "apple" describe general attributes, so they are excluded. "Core" and "Cider" are unrelated to "brown apple." "Withered" relates to a decayed state, which applies to "brown apple" but not to "brown" or "apple" alone. To interpret "withered" naturally, we select "unappetizing" as the emergent property. So the answer is \{\{"property": "unappetizing"\}\}\\
\\
Above answer is correct because property "unappetizing" does not belong to either "brown" and "apple", but belong to brown apple\\
\\
<Example 2>\\
- Combination: burned banknote\\
- Relevant concepts: ['paper', 'ash', 'money', 'value']\\
- Wrong answer: The goal is to find an emergent property of a “burned banknote” that does not exist in either “burned” or “banknote” individually. The attributes “paper” and “money” describe general properties of a banknote, so they are excluded.Likewise, “ash” describes a general property of something that is burned, so it is also excluded. While a banknote possesses “value,” this characteristic disappears once the banknote is burned. Consequently, to convey the idea of “valueless” naturally, we choose “useless” as the emergent property. So the answer is is: \{\{"property": "useless"\}\}\\
\\
Above answer is wrong because "burned" something can directly belong to property "useless". Ensure that the emergent property you generate does not directly describe any of the individual words but is a characteristic of the combination as a whole.\\
\\
Then let's begin:\\
\bottomrule
\end{tabular}
    \caption{Task instruction for Spread Activation Method in Property Induction.}
    \label{tab:pi_prompt_sa}
\end{table*}
\begin{table*}[htbp]
\scriptsize
\centering
\begin{tabular}{@{}p{\linewidth}@{}}
\toprule
\textbf{Task Instruction for Property Type Prediction}\\
\midrule
Instructions:\\
1. You are given a combination and property. Your task is to predict a type of property.\\
2. Definition of each property type is as follows:\\
    - Emergent: The property emerges from the combination of components.\\
    - Component: The property is inherited by component of the combination.\\
    - Canceled: The property is canceled out by the combination of components.\\
    - Others: The property is not related to the combination nor components.\\
3. Use the previous examples to learn the task.\\
4. Answer in dictionary format: \{\{"property\_type": "\{\{property\_type\}\}"\}\}. Do not include other formatting.\\
    \\
<Example 1>\\
- Combination: peeled apple\\
- Property: round\\
- Correct answer: \{\{"property\_type": "component"\}\}\\
\\
Above answer is correct because property "round" is inherited by component "apple".\\
\\
<Example 2>\\
- Combination: burned banknote\\
- Property: useless\\
- Wrong answer: \{\{"property\_type": "emergent"\}\}\\
\\
Above answer is wrong because modifier "burned" directly elicit property "useless".\\
\\
\\
Then let's begin:\\
\bottomrule
\end{tabular}
    \caption{Task instruction for property type prediction.}
    \label{tab:pt_prompt}
\end{table*}
\begin{table*}[htbp]
\scriptsize
\centering
\begin{tabular}{@{}p{\linewidth}@{}}
\toprule
\textbf{Prompt for LLM-as-a-Judge}\\
\midrule
Concepts are characterized by properties. For example, the concept "a chicken in front of a fox" strongly exhibits the property "in danger." When given a concept and a property, your task is to evaluate how much the concept has the property on a scale from 1 to 10. You should follow the format: \{\{"relevance": your\_relevance\_score\}\}\\
\\
Use the following scoring criteria to assign a relevance score:\\
- \{\{"relevance": 1\}\}: The concept does not have the property at all.\\
- \{\{"relevance": 2-3\}\}: The concept rarely has the property.\\
- \{\{"relevance": 4-6\}\}: The concept sometimes has the property.\\
- \{\{"relevance": 7-8\}\}: The concept usually has the property, but not always.\\
- \{\{"relevance": 9\}\}: The concept almost always has the property.\\
- \{\{"relevance": 10\}\}: The concept always has the property.\\
---\\
Examples:\\
---\\
Concept: Rusty\\
Property: Useless\\
Relevance: \{\{"relevance": 7\}\}\\
---\\
Concept: A chicken in the cage\\
Property: In danger\\
Relevance: \{\{"relevance": 2\}\}\\
---\\
Concept: A chicken in front of a fox\\
Property: In danger\\
Relevance: \{\{"relevance": 9\}\}\\
---\\
Concept: Wrench\\
Property: Destructive\\
Relevance:\\
\bottomrule
\end{tabular}
    \caption{Prompt for LLM-as-a-judge.}
    \label{tab:prompt_judge}
\end{table*}


\subsection{Collecting human solutions}
\label{subsec:collection_human_solutions}
The students reported to need 25 to 45 minutes per 30 questions. We paid \$78 per 167-question test sheet and \$94 for 200-question test sheet; This exceeds the minimum wage in the country the authors and students are located in. The question format is as below.

\textbf{(Property induction - emergent property)}
\\
Conceptual combination: \textbf{`\{noun phrase\}'}, Type-of-property: \textbf{`emergent property'}
\\
\textbf{Q.} What is a property of \textbf{`{\color{pink}\{noun phrase\}}'} that \textbf{`{\color{softsky}\{head noun\}}'} or \textbf{`{\color{softsky}\{modifier\}}'} doesn't have? \\
\textbf{A:} \rule{5cm}{0.15mm}

\vspace{1em}

\textbf{(Property induction - canceled property)}
\\
Conceptual combination: \textbf{`\{noun phrase\}'}, Type-of-property: \textbf{`canceled property'}
\\
\textbf{Q.} What is a canceled property which is a property that belongs to \textbf{`{\color{pink}\{head noun\}}'} or \textbf{`{\color{pink}\{modifier\}}'} but does \textbf{not} belong to a noun phrase \textbf{`{\color{softsky}\{noun phrase\}}'}? \\
\textbf{A:} \rule{5cm}{0.15mm}

\vspace{1em}

\textbf{(Noun phrase completion)}
\\
Head noun: \textbf{`\{head noun\}'}, Property: \textbf{`\{property\}'}
\\
\textbf{Q.} What is a noun phrase using \textbf{`{\color{pink}\{root\}}'} to be \textbf{`{\color{pink}\{property\}}'}? \\
\textbf{A:} \rule{5cm}{0.15mm}

\begin{figure*}[t!]
\centering
\includegraphics[width=\linewidth]{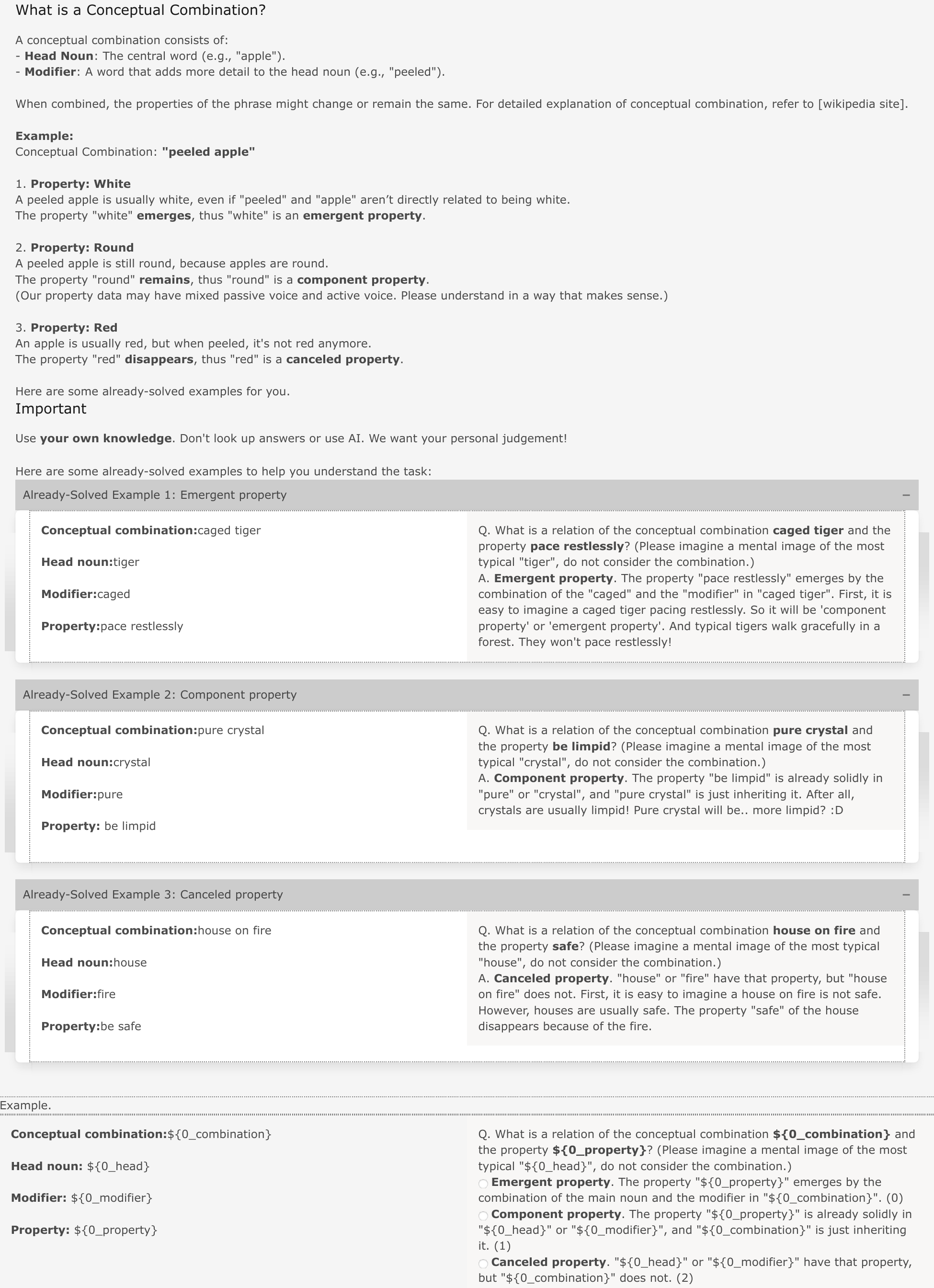}
\caption{Instructions provided for raters in Amazon Mechanical Turk to collect human annotations for data quality.}
\label{fig:mturk_type_tagging_instruction}
\end{figure*}

\subsection{Recruiting Participants for Property Type Classification}
\label{subsec:property_type_mturk_detail}

To check human ability for property type prediction in Section~\ref{subsec:classification_task_result}, we recruited capable test participants who did \textbf{not} participate by Amazon Mechanical Turk (MTurk).

We provided the participants with the conceptual combination and property and asked them to classify the property type. The definition of the conceptual combination and the property types are described in the instructions. Three annotated data instances (one per each property type) are also provided. Instruction is shown in Figure~\ref{fig:mturk_type_tagging_instruction}. As a result, when we regard the majority vote as a human-annotated label, we get an accuracy of 87\% for the emergent property, 79\% for the component property, and 76\% for the canceled property, with a total accuracy of 81\%.

We prepared the qualification Human Intelligence Task (HIT), comprised of 10 thoroughly verified examples with a payment of \$1.0. We recruited participants from AU, CA, NZ, US, and GB, with more than 10000 HITs approved, and a HIT approval rate greater than 98\%. Among 40, this process resulted in 12 participants.

After qualification, we asked raters with a payment of \$0.2 per HIT. Each example was evaluated by three annotators and the inter-annotator agreement was 0.59 in Fleiss' Kappa~\cite{fleiss1971measuring}.

\subsection{Evaluation of relation between LLM-as-a-Judge and Human Judge}
\label{subsec:evaluation_between_llm_as_a_judge_and_human_judge}

\begin{figure*}[t!]
\centering
\includegraphics[width=\linewidth]{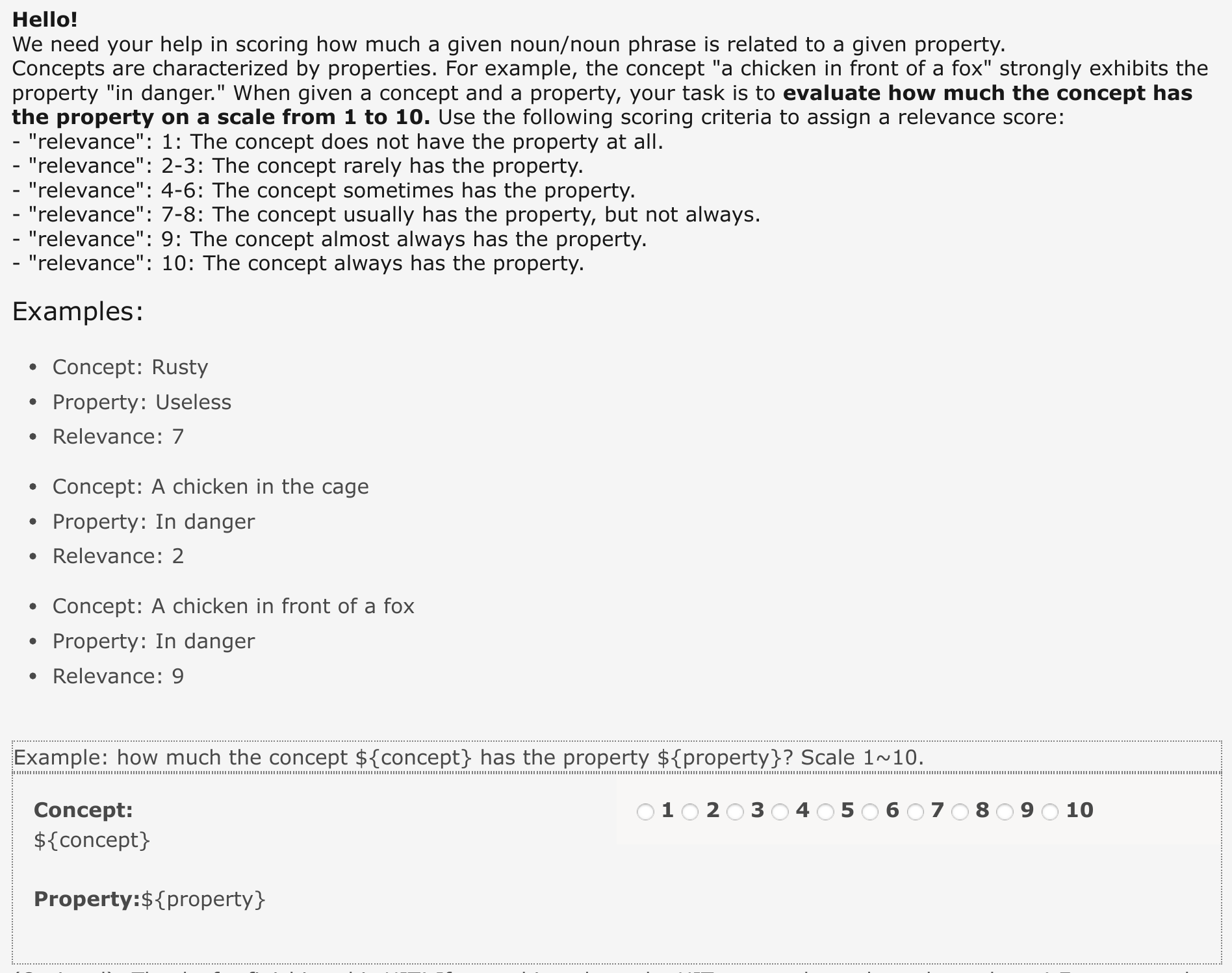}
\caption{Instructions provided for raters in Amazon Mechanical Turk to collect the relevance score between the given property and a concept.}
\label{fig:mturk_relevance_tagging_instruction}
\end{figure*}

To ensure the quality of evaluation metric in \mydata, we measure a correlation between LLM-as-a-judge and human ratings in Section~\ref{subsec:relevance_llm_human}. Like the previous subsection, we hire capable raters in Amazon Mechanical Turk (MTurk) who did \textbf{not} participate in the data annotation process before this test.

The basic qualifications are also made by nationality (AU, CA, NZ, US, and GB), the number of HITs approved (10000), and the HIT approval rate (greater than 98\%). We performed a qualification test with a payment of \$0.1 and chose 12 raters among the 85 applicants who had completed more than 5 qualification HITs. We paid \$0.1 for the main rating. Refer to Figure~\ref{fig:mturk_relevance_tagging_instruction} for the instruction that we used for the qualification and the main judge task.
\end{document}